\documentclass[lettersize,journal]{IEEEtran}
\usepackage{amsmath,amsfonts}
\usepackage{algorithmic}
\usepackage{algorithm}
\usepackage{array}
\usepackage{textcomp}
\usepackage{stfloats}
\usepackage{url}
\usepackage{verbatim}
\usepackage{graphicx}
\usepackage{siunitx}
\usepackage{subcaption}
\usepackage{cite}

\usepackage{xcolor}
\usepackage{amsmath}

\usepackage{amssymb}
\usepackage{bm}
\usepackage{svg}
\usepackage{hyperref}
\usepackage{cite}
\usepackage[normalem]{ulem}
\usepackage{bbm}

\usepackage{longtable}
\usepackage{booktabs} 

\newif\ifincludeappendix 
\includeappendixfalse

\newcommand{\algname}{S.S. Explorer }
\newcommand{\algnamenospace}{S.S. Explorer}

\newcommand{\immIndicator}{{i}}

\newcommand{\safeStochArrivalIndicator}{{m}}
\newcommand{\immOperator}{{I}}
\newcommand{\safeStochReturnOperator}{{\bar{R}}}
\newcommand{\safeStochArrivalOperator}{\bar{m}}
\newcommand{\state}{{x}}
\newcommand{\randomState}{\mathbf{{\state}}}
\newcommand{\stateSpace}{{X}}

\newcommand{\action}{{a}}
\newcommand{\actionSpace}{{A}}
\newcommand{\transitionFunction}{{T}}
\newcommand{\performanceFunction}{{f}}
\newcommand{\heuristicFunction}{{h}}
\newcommand{\safetyFunction}{{q}}

\newcommand{\safetyThreshold}{{q_{T}}}

\newcommand{\safeSet}{{S}}

\newcommand{\agent}{{agent}}
\newcommand{\iter}{{t}}
\newcommand{\agentspace}{{agent }}

\newcommand\nsnote[1]{\textcolor{red}{[NS: #1]}}

\usepackage{amsthm}
\newtheorem{definition}{Definition}

\title{Safe Stochastic Explorer: Enabling Safe Goal Driven Exploration in Stochastic Environments and Safe Interaction with Unknown Objects}
\author{
Nikhil Uday Shinde$^{1}$, Dylan Hirsch$^{1}$, Michael C. Yip$^{1}$, Sylvia Herbert$^{1}$\\
$^{1}$University of California San Diego\\
{\tt\small \{nshinde, dhirsch, yip, sherbert\}@ucsd.edu}
}

\begin{document}

\maketitle

\begin{abstract}
Autonomous robots operating in unstructured, safety-critical environments, from planetary exploration to warehouses and homes, must learn to safely navigate and interact with their surroundings despite limited prior knowledge. 
Current methods for safe control, such as Hamilton-Jacobi Reachability and Control Barrier Functions, assume known system dynamics. 
Meanwhile existing safe exploration techniques often fail to account for the unavoidable stochasticity inherent when operating in unknown real world environments, such as an exploratory rover skidding over an unseen surface or a household robot pushing around unmapped objects in a pantry. 
To address this critical gap, we propose Safe Stochastic Explorer (\algnamenospace) a novel framework for safe, goal-driven exploration under stochastic dynamics. 
Our approach strategically balances safety and information gathering to reduce uncertainty about safety in the unknown environment. 
We employ Gaussian Processes to learn the unknown safety function online, leveraging their predictive uncertainty to guide information-gathering actions and provide probabilistic bounds on safety violations. 
We first present our method for discrete state space environments and then introduce a scalable relaxation to effectively extend this approach to continuous state spaces. 
Finally we demonstrate how this framework can be naturally applied to ensure safe physical interaction with multiple unknown objects. 
Extensive validation in simulation and demonstrative hardware experiments showcase the efficacy of our method, representing a step forward toward enabling reliable widespread robot autonomy in complex, uncertain environments. 
\end{abstract}

\section{Introduction}
Autonomous robots are increasingly expected to operate in unstructured, safety-critical environments, where complete prior knowledge of the surroundings cannot be assumed. 
Examples include planetary exploration missions, such as a Mars Rover navigating through unfamiliar, potentially hazardous terrain, and disaster response scenarios where robots must make informed decisions to avoid unsafe regions that could lead to failure or harm humans. 
While such high-stakes applications highlight the importance of safety, the need extends even further to far more common settings.
In everyday environments such as warehouses, retail stores, hospitals, and even people's homes, robots encounter countless unknown objects and dynamic conditions. 
Here, safety includes preventing damage to fragile items and avoiding hazardous spills that can create unsafe conditions for human collaborators. 

The ability to safely and autonomously interact with unknown environments is therefore critical to realizing the societal benefits of robotics at scale. 
Robots that can reason about safety while exploring can unlock transformative applications: providing home assistance and mobility support for elderly individuals and people with disabilities; reducing the physical strain of repetitive tasks such as warehouse shelf-stocking; and removing the need for humans to enter dangerous or inaccessible areas.
However, achieving this vision requires methods that allow robots not only to accomplish tasks in unknown environments but also to manage uncertainty and remain within safe operating bounds at all times.

While there has been extensive work on robotic safety, existing approaches leave critical gaps.
Hamilton–Jacobi reachability (HJR) \cite{HJR} methods compute robust safe sets but assume known dynamics and safety functions to specify their failure sets. 
Control barrier functions (CBFs) \cite{CBFAmes} act as minimally invasive safety filters but similarly require accurate system models and can often be hard to synthesize. 
Learning-based approaches have achieved impressive performance in manipulation and non-prehensile interaction, yet they rarely actively consider safety and ensure safety with high confidence. 
Prior safe exploration methods, such as those based on Gaussian process optimization or safe reinforcement learning, reason about unknown environments \cite{safeOPT, safeIML} but typically neglect stochasticity in the dynamics. 
In real-world settings, however, uncertainty is unavoidable: perfect modeling is infeasible, and safety frameworks must explicitly account for this.

In this work, we focus on the problem of safe goal-driven exploration in unknown environments under stochastic dynamics. 
We consider a robot tasked with completing an objective in an environment whose dynamics and safety characteristics are not fully known and where transitions are inherently uncertain. 
Such uncertainty can arise from factors like traversing unknown or unstable terrain—e.g., a rover skidding on sand or ice—or from interacting with objects whose physical properties are unknown and whose motion upon contact are difficult to predict. 
The robot must balance two competing needs: (i) exploring and gathering information to reduce uncertainty about the environment and its safety properties, and (ii) ensuring with high confidence that it remains within a safe set throughout its operation. 

To address these challenges, we propose Safe Stochastic Explorer (\algnamenospace), a novel framework for safe goal-driven exploration in unknown environments that explicitly accounts for stochastic dynamics. 
Our major contributions are as follows:
\begin{enumerate}
    \item We introduce a method for safe exploration in discrete state spaces with stochastic transitions
    to ensure 
    that the \agentspace remains within a safe set. 
    The approach uses Gaussian processes to model the unknown safety function and leverages the learned uncertainty and known dynamic stochasticity to guide information-gathering actions while bounding the probability of safety violations.
    \item We extend our framework to continuous state spaces, developing a scalable relaxation that retains safe exploration while improving efficiency for continuous state space, real-world domains.
    \item We demonstrate how this formulation can be naturally augmented to ensure safe physical interaction with multiple unknown objects, a crucial capability for household and warehouse robots that must interact with items of initially unknown mass, stability, or frictional properties.
\end{enumerate}
We validate our approach extensively in simulation and showcase real-world experiments on a manipulator performing safe interaction tasks. 
The results highlight the potential of safety-aware exploration under uncertainty to enable broader and safer integration of autonomous robots in unstructured environments. 

The remainder of this paper is organized as follows.
Section II reviews related work in robotic safety and safe exploration. 
Section III presents our method, detailing the theoretically grounded discrete-state formulation, its scalable continuous-state relaxation, and its extension to safe object interaction tasks. 
Section IV describes our experimental setup, and reports results from both simulation and real-world hardware experiments. 
Finally, Section V discusses limitations, and future directions of our work.

\section{Related Works}




Ensuring safety is a key thrust of research in robotics, yet it remains a significant challenge for robots operating in unstructured environments with partially known dynamics. 

\subsection{Generating Safe Control Policies for Stochastic Dynamics in a priori Unknown Environments}

Traditional safety analysis methods offer strong theoretical guarantees but rely on a complete characterization of a system and its safety, posing a significant challenge for online adaptation in unknown environments. 
One such method, Hamilton-Jacobi Reachability (HJR), provides a principled framework for safety by solving for a value function that defines a control invariant safe set and optimal control policy to remain safe \cite{HJR}.
Computing this value function, however, relies on accurate system models and predefined characterizations of disturbance sets and environment properties. 
Furthermore, classical dynamic programming (DP) based approaches to solving for the value function are restricted to lower-dimensional systems due to the 'curse of dimensionality' \cite{HJR}. 
Although recent advancements using deep learning and physics informed neural networks has enabled learning the safety value function for higher dimensional systems \cite{deepreach}, the requirement for precise models and significant computation time to learn a new function makes HJR approaches generally ill suited for online adaptation to unknown scenarios. 
Some efforts explore extending HJR approaches for online adaptation by parameterizing the value function based on specific unknowns, such as disturbance bounds, allowing for online inference from pre-computed or learned value functions \cite{parameterizedHJR, spacetotime}; yet, this requires both offline specification and precomputation across a range for a specified parameter, as well as real time estimation capabilities of that parameter. 
Other approaches like Neural Operators \cite{mochenNeuralOperator} can generate value functions for different obstacle maps, but remain specific to the dynamics and disturbance bounds used during training. 
These challenges become more computationally expensive when faced with system stochasticity \cite{margaretStochasticHJR, inkyuStochasticHJR}.

Control Barrier Functions (CBFs) have also gained popularity with their use as minimally invasive safety filters. 
Much like the HJR value function, a CBF is a continuously differentiable function whose value describes a system's level of safety and whose gradients inform an online QP based controller on when to override a nominal policy and preserve safety \cite{CBFAmes}. 
However, synthesizing a valid CBF is challenging, demanding intimate knowledge of the system dynamics and operating environment. 
Many approaches simplify synthesis by focusing on simple systems or using reduced order models. 
Other methods that use HJR to synthesize viscosity-based CBFs \cite{backtobase, spacetotime} carry the limitations of the underlying HJR methods. 
Learning based approaches such as \cite{NCBFsimin, sncbf} still depend on simulators or high fidelity models to generate training data. 
Ultimately all these approaches require  intimate system knowledge making CBFs difficult to adapt online to unknown situations. 
Work on updating CBFs in unknown environments is thus typically forced to rely on assumptions of simplified dynamics \cite{kehanCBF, gilbertCBF} or potentially cumbersome grid-based dynamic programming approaches \cite{refineCBF}. 

A common thread across these classical and learning-assisted methods is their reliance on analytical models. 
To address the limitations of model-based methods, the field of Safe Reinforcement Learning (Safe RL) aims to learn high-performing policies that also satisfy safety constraints. 
Many approaches formulate this as a constrained Markov Decision Process (CMDP), incorporating safety-related costs into the learning objective to produce a performant, yet safe final policy \cite{CPO}. 
Other works have drawn inspiration from HJR, changing the losses and policy updates with modified Bellman equations to better learn safety-aware value functions for complex systems \cite{RCPO, RCPPO, DOHJPPO}. 
However, these approaches do not guarantee safety during the learning process itself. 
To address this, another line of work in safe RL focuses on safe exploration, where the agent must collect data without violating safety constraints. 
This is often done by adding minimally invasive safety filters into the RL loop to ensure every action in the real world is safe \cite{backtobase}. 
While effective, all these methods depend on a relatively accurate simulator to generate safe and performant policies. 

Beyond the context of RL, Safe Exploration is a large field of research focused on the problem of optimizing how an agent can efficiently sample and learn about a system or its environment while respecting safety critical constraints. 
Seminal work in this area introduced Bayesian optimization-based methods, like SAFEOPT 
\cite{safeOPT, safeOPT2}, which guarantees that safety constraints are not violated during exploration. 
Subsequent work extended this to physical robotic systems by adding Markovian constraints (SAFEMDP)  
\cite{safeMDP}
, restricting the agent to only sample an environment in a markovian manner similar to the restrictions on a physical system. 
This was later extended to focus on the problem of goal-directed exploration in the GoOSE algorithm proposed in  \cite{safeIML}.
These methods provide a powerful framework for guaranteed safe-learning in unknown, unexplored environments. 
However, these methods fail to properly account for the stochasticity that characterizes state transitions in the real world. 
Our work in this paper bridges that gap. 
We build directly upon the principles of goal-driven safe exploration presented in GoOSE, proposing a novel algorithm that explicitly handles stochastic transitions. 
We first present our algorithm for discrete state space cases, then introduce relaxations to extend it to continuous state spaces before extending it to the problem of safe physical interaction with multiple unknown objects.  

\subsection{Safe Physical Interaction}
Most classical and learning-assisted methods for generating safe control policies focus primarily on collision avoidance as the definition of safety. 
This overlooks a vast class of robotic tasks where physical interaction is not only desired but necessary. 
In this paper we consider the problem of completing a task where we are forced to safely interact with unknown movable objects. 
There is a rich landscape of literature focusing on object interaction in the context of performant policies for object manipulation tasks. 
These methods achieve performant prehensile and non-prehensile object manipulation using both classical and analytical model-based methods  
\cite{nonprehensile-object-transport}
and learning based approaches like in 
\cite{lowrey2018reinforcementlearningnonprehensilemanipulation, zhang2023reinforcementlearningbasedpushing}
. 
Works like 
\cite{oller2024tactiledrivennonprehensileobjectmanipulation}
explore additional modalities like tactile sensing to improve performance for manipulation. 
While other works use human-in-the-loop approaches to handle complex scenarios like cluttered environments 
\cite{papallas2020nonprehensilemanipulationclutterhumanintheloop}. 
Yet the aforementioned works fail to explicitly consider safety. 

While recent works have begun to consider safety during manipulation many of these approaches still do not consider safety during contact, with those that do consider contact requiring rich datasets or accurate simulations. 
In works like 
\cite{mpcDLO}
, though they safely manipulate a complex deformable object, safety is considered by simply extending collision avoidance from just the robot to also consider the manipulated object. 
Other works such as 
\cite{safePicking}
, for rigid object manipulation,  and 
\cite{JIGGLE}
, for deformable manipulation, fail to explicitly handle constraints on safety, instead treating it only as a soft constraint via reward penalties when learning or optimizing for the actions to be taken. 
Works like 
\cite{nakamura2025generalizingsafetycollisionavoidancelatentspace}
have begun to explicitly reason about safety by leveraging learned latent spaces to create safety filters to ensure safety during complex object interactions. 
Yet these methods too suffer from requiring either a high-quality simulator or multiple datapoints to be collected while interacting with the desired object to learn about the safety of the latent space. 

The challenges of safe object interaction are compounded when interacting with unknown objects. 
Works that focus on the manipulation of unknown objects, often formulate the problem as perception and modeling problems. 
These works update the geometric models of unknown objects to enable task completion  
\cite{mitash2020taskdrivenperceptionmanipulationconstrained, SimNet}
, but often fail to consider non-visual properties like mass or friction. 
Other approaches focus explicitly on inferring these hidden properties through interaction. 
These works extend across rigid objects 
\cite{shinde2023objectcentricrepresentationsinteractiveonline, 7280802, adaptigraph, allisonPalpating}
or even dynamics for tasks like human-robot collaboration 
\cite{7989204}. 
While these methods successfully learn models for performance, they do not explicitly reason about the safety of the interactions used to gather this information. 
In our work we highlight the problem of completing a task while ensuring safe interaction with unknown objects. 
We demonstrate how this problem can be formulated and solved within the framework of safe goal-driven exploration.

\section{Problem Statement}
In this work we will solve $3$ distinct variations of the problem of safe goal-driven interaction in unknown stochastic environments. 
We will start the methods section by defining the overarching general problem statement.
We will then discuss individual variations of the problem statement, beginning with two general variations referred to as the Discrete and Continuous Cases, followed by an application-specific case that we refer to as the object case. 


We consider an \agentspace operating in a state space $\state \in \stateSpace$. 
From any state $\state$ the \agentspace can take actions from a state-dependent discrete action set $\action \in \actionSpace(\state) \subseteq \actionSpace$. 
An action taken from state $\state$ at time $t$ results in a stochastic transition to a random variable next state $\randomState_{t+1}$ according to: 
\begin{equation}
\randomState_{t+1} = \transitionFunction(\state_{t}, \action_{t}) \sim p(\state_{t+1}|\state_{t}, \action_{t}). 
\end{equation}
Here $\transitionFunction(\state_{t}, \action_{t})$ is the transition function, and the transition probabilities $p(\state_{t+1}|\state_{t}, \action_{t})$ are modeled by a known probability distribution over possible arrival states. 
The environment is governed by two functions that are unknown a priori:

\noindent \textbf{1)} The \textbf{Safety Function}, $\safetyFunction(\state)$ - A state, $\state$, is considered unsafe if $\safetyFunction(\state) > \safetyThreshold$, where $\safetyThreshold$ is a known safety threshold.

\noindent \textbf{\textbf{2)}} The \textbf{Performance Function}, $\performanceFunction(\state)$ – This task-specific function quantifies the desirability of a state.
For example, in a mining task it could represent the amount of gold at $\state$, whereas in a Mars exploration task (as in \cite{safeIML}), it can represent the altitude of a state, where higher altitudes offer better surveillance.

The \agent’s goal is to explore the environment and reach a final state $\state_{T}$ that maximizes the performance function while avoiding safety violations: 
\begin{equation}
\begin{aligned}
\arg\max_{\action_{0} \dots \action_{T-1}} \qquad &\performanceFunction(\state_{T}) \\
\text{s.t.} \qquad &\safetyFunction(\state_{t}) \leq \safetyThreshold~~ \forall t \in \{0,1,\dots,T\}.
\end{aligned}
\end{equation}

While the true performance function $\performanceFunction$ is a priori unknown, we assume the \agentspace is equipped with a heuristic function $\heuristicFunction(\state)$ that provides guidance during exploration.
In line with \cite{safeIML}, rather than focusing on unsafe exploration itself, which is centered around understanding and maximizing $\performanceFunction(\state)$, we assume access to an existing unsafe exploration algorithm that outputs intermediate task objectives, specific states $x^*$, or desired goal regions to be sampled. 
Our work addresses this complementary challenge on how to navigate and explore so as to safely and successfully reach and sample the desired points or goal regions provided by the unsafe exploration algorithm, prioritizing understanding $\safetyFunction(\state)$ to reach $x^*$ without violating $\safetyThreshold$. 
Thus, we introduce the general safe exploration problem formulation that demonstrates how our work can be used for a broad range of tasks, while we ground our study in the concrete setting of safely exploring the environment to reach designated goal regions or target sample points. 

We now introduce the assumptions required to make the problem statement tractable. 
We assume that the state space admits a positive definite distance kernel, $k(\cdot, \cdot)$, and that the safety function, $\safetyFunction$, has a bounded norm in the corresponding Reproducing Kernel Hilbert Space (RKHS) \cite{learningwithkernels}
: $||\safetyFunction||_{k} \leq L$. 
This bounded norm condition imposes a continuity requirement on $\safetyFunction$, ensuring that $\safetyFunction$ is $L$–Lipschitz continuous with respect to the associated distance metric: $d(x, z)=\sqrt{k(x, x) - 2k(x, z) + k(z, z) }$, i.e.
\begin{align}
    |\safetyFunction(x) - \safetyFunction(z)| \leq Ld(x,z) \quad \forall x,z \in \stateSpace
\end{align}
In addition to these assumptions on regularity, we assume that the \agentspace is initialized within a control invariant starting safe set, $\safeSet_{0}$, 
where it can always take an action that will surely not violate safety. 
Finally, we assume that the \agentspace has full state observability and the ability to continually re-plan its trajectory as new information becomes available.

In the upcoming sections we will define 3 particular subsets of the general problem discussed above, along with 3 variants of our method, \algname,  proposed to solve them.

\section{Framework: Discrete Formulation}
We begin by defining the problem statement for what we will refer to as the discrete case. 
Here we consider the general problem statement, where $\stateSpace$ is a finite, discrete state space, and the stochastic transition function, $\transitionFunction$, is modeled using a discrete probability distribution, $p(\state_{t+1}|\state_{t}, \action_{t})$. 

\textbf{\algname Discrete Case: }
To discuss the solution to the discrete case we begin by posing the \agent's motion through the state space as a Markov chain. 
Given some action sequence $\mathbf{\action}  = (\action_{0}, \action_{1}, \action_{2}, \dots), \action_{t} \in \actionSpace$ and initial state $\state \in \stateSpace$, we define $\mathbf{X}_{\state}^{\mathbf{\action}}(\cdot)$ to be the Markov chain given by
\begin{align}
    &\mathbf{X}_{\state}^{\mathbf{\action}}(t+1) = f(\mathbf{X}_{\state}^{\mathbf{\action}}(t), \action_{t}) \\
    &\mathbb{P}\left(\mathbf{X}_{\state}^{\mathbf{\action}}(0) = \state\right) = 1 \nonumber 
\end{align}
Though the actions available are state dependent, i.e. $\action \in \actionSpace(\state)$, we modify the transition function to result in an inaction for any invalid choice of actions: $f(\state_{t}, \action_{t}) = \state_{t} \: \forall \action_{t} \notin \actionSpace(\state)$. 
With this augmentation we can phrase our system given a set of actions as a Markov chain without loss of generality. 

We solve the above problem statement using a set-based solution. 
Starting with the initial safe set, we propose an algorithm to grow the safe set and navigate through the space to achieve the objective. 
We begin by defining safety at a conceptual level. 
A state is considered \textbf{completely} safe if it satisfies the following: 
\begin{itemize}
    \item \textbf{Immediate Safety: } A state $\state$ is immediately safe if its safety cost $\safetyFunction(\state)$ does not exceed the safety threshold: $\safetyFunction(\state) \leq \safetyThreshold$.
    \item \textbf{Safe Stochastic Return: } A state is stochastic-returnably safe if there is a guaranteed trajectory back to the initial safe set without exiting the set of Immediately Safe States. 
    This ensures recursive feasibility: that the \agentspace does not get stuck, and can continue safe exploration. 
    \item \textbf{Safe Stochastic Arrival: } 
    A state, $\state$, is stochastic-arrivably safe if the \agentspace can transition to that state from the current safe set while ensuring that the \agentspace does not exit the set of states that satisfy immediate safety and safe stochastic return. 
\end{itemize}
While we outline these conceptual definitions of safety here, we will formalize these definitions as we discuss the operators that we will use to construct these sets. 

To solve the discrete case, we expand the set of completely safe states that satisfy these $3$ safety criteria through iterative set expansion and refinement operators. 
At each expansion iteration, we begin by using our current knowledge of the safety function at sampled states to expand the immediate safe set. 
This gives us a larger set of states that we believe to be immediately safe: $\{\state \in \stateSpace | \safetyFunction(\state) \leq \safetyThreshold\}$.
We then refine this set to the states which satisfy the remaining safety criteria, first restricting to those that are also stochastic-returnably safe, with further refinement to those that are also stochastic-arrivably safe. 
These $3$ steps of expansion followed by refinement give us the set of completely safe states. 
In this section we will rigorously define these operators, followed by an explanation on how to use them for safe goal-driven exploration.

We begin our discussion by defining the $2$ refinement operators: the Safe Stochastic Return Refinement Operator, $\safeStochReturnOperator$, and Safe Stochastic Arrival Refinement Operator, $\safeStochArrivalOperator$. 
To do this we will first assume that at each expansion iteration we have access to an oracle that will give us an expanded set of states, $\immOperator(S_{t})$, that are immediately safe which are a super set of our currently completely safe states, $S_{t}$. 
We will first focus on the refinement operators used to restrict this set $\immOperator(S_{t})$, after which we will remove the oracle assumption and define the true Immediate Safe Expansion Operator, $\immOperator$.

\subsection{Stochastic Return Refinement Operator} We begin by introducing the refinement operator that ensures the system can return to the completely safe set. 

\begin{definition}[Stochastic Return Refinement Operator, $\safeStochReturnOperator$] 
Given sets $S, I \subseteq \stateSpace$ with $S$ control invariant and $S \subseteq I$, let $\safeStochReturnOperator(S, I)$ be the largest control invariant set in $I$ such that $S \subseteq R(S, I$ and for each $\state \in \safeStochReturnOperator(S, I)$ there exists an action sequence $\mathbf{\action} = \{\action_{0}, \action_{1}, \dots \}$ satisfying
\begin{align}
    \mathbb{P}( \exists t \in \mathbb{N}_{0} ~ \mathbf{X}^{\mathbf{a}}_{x}(t) \in S) = 1.
\end{align}
\end{definition}
This operator thus outputs a control invariant set of states such that there always exists a (potentially infinitely long) set of actions that enable the \agentspace to return back to the set $S$ in spite of stochasticity. 

We now discuss how to construct this control-invariant set in practice.
To build up to the operator, $\safeStochReturnOperator$, we first begin by considering how one would define the analogous operator in an environment with deterministic transitions. 
We start by defining $r_{\text{det}}^1$, the deterministic single-step return operator. 
The set $r_{\text{det}}^1(S, I)$ restricts $I$ to the set of states that are at most $1$ step away from returning to the set $S$: 
\begin{align}\label{eq:r_det}
    r_{\text{det}}^1(S, I) = \{ x \in I | \exists  a \in A(x) ~ T_{\text{det}}(x, \action) \in S\} \cup \{S\} \nonumber
\end{align}
where $T_{\text{det}}$ is a deterministic transition function. 
To find the set of states that have an $n$ step deterministic path back to the set $S$ we can simply recurse this operator $n$ times. 
\begin{equation} \label{eq:r_det_recursion}
    r_{\text{det}}^n(S, I) = r_{\text{det}}^1(r^{n-1}_{\text{det}}(S, I), I)
\end{equation}

Returning to the stochastic case in which we are interested, this notion of single-step return is not as straightforward. 
Consider evaluating whether a state $x$ satisfies the criteria of safe stochastic return. 
While some part of the state distribution after a transition from state $x$ may return to a desired set, $S$, part of the distribution may fail to do so. 
However, if these states temporarily arrive outside $S$, but still have a path back to $S$, we must still consider state $x$ to satisfy safe stochastic return. 
While this return path could be arbitrarily long in principle, in our stochastic setting with non-zero probability of return, the expected first arrival time to the desired return set $S$ is finite. 
We now discuss how one can compute the set $\safeStochReturnOperator(S, I)$ in practice. 

To construct $\safeStochReturnOperator(S, I)$ in our stochastic setting, we start by defining the single-step safe stochastic return operator $r$. 
\begin{align}
    r^1(S, I) &= \{ x \in I | \exists  \action \in \actionSpace(x) \text{ s.t. }\\
    & \mathbb{P}(T(x, a) \in S) > 0, \mathbb{P}(T(x, a) \in I) = 1\} \nonumber \cup \{S\}\nonumber \\
    r^n(S, I) &= r^1(r^{n-1}(S, I), I) \nonumber
\end{align}
The single-step operator $r^1(S, I)$ restricts the set $I$ to only those states in $I$ such that there is some probability of returning to set $S$ with action $a \in \actionSpace(x)$ without any chance of exiting set $I$. 
Because the state space is finite, the $n$-step operator $r^n(S, I)$ is guaranteed to converge within $|I|$ steps. 
We let $\bar{r}(S, I) = 
r^{|I|}(S, I)$. 

The set $\bar{r}(S, I)$ thus consists of all the states, $x \in I$, from which one can potentially return to the desired set $S$ without risking exit from the set $I$. 
This however, does not guarantee the \agent's ability to return to $S$, but only ensures containment within $I$, when derailed from its path due to one stochastic transition.  
To certify return to $S$, we need to further restrict the computed set $\bar{r}(S, I)$, to ensure that after any one stochastic the \agentspace is still contained within $\bar{r}(S, I)$. This would guarantee the ability to chart a deterministic return path back to set $S$,  capable of withstanding one stochastic transition. 
To protect against additional stochastic transitions, we can further recurse this restriction to compute the control invariant set $\safeStochReturnOperator(S, I)$. 
This set ensures that there is always a safe path back to $S$ in spite of stochastic transitions that may deviate from a deterministically planned path. 
This gives us the final Safe Stochastic Return Operator, $\safeStochReturnOperator$: 
\begin{align}\label{eq:safe_stoch_return_operator}
    R^{1}(S, I) &= \bar{r}(S, I) \nonumber \\
    R^{n}(S, I) &= \bar{r}(S, R^{n-1}(S, I)) 
\end{align}
We then compute $\bar{R}(S, I)$ as follows: 
\begin{align}
    \bar{R}(S, I) = \lim_{n \rightarrow |I|} R^{n}(S, I).
\end{align}
 As $R^{n}$ is a restriction on finite discrete set $I$, $\bar{R}$ converges in finite steps when $R^{n} = R^{n-1}$.

In expansion iteration $t+1$, we compute the set $\safeStochReturnOperator(S_{t}, \immOperator(S_{t}))$ where $S_{t}$ is the previous completely safe set, and $\immOperator(S_{t})$ is the immediately safe expansion given by the oracle. 
$\safeStochReturnOperator(S_{t}, \immOperator(S_{t}))$ provides the set of immediately safe states from which the system can return to the previous completely safe set, $S_{t}$, guaranteeing task continuation by preventing the \agentspace from getting stuck.



\subsection{Stochastic Arrival Refinement Operator}
When planning to sample a new target state outside of $S_{t}$, we must not only consider whether the single target state is immediately and stochastic-returnably safe, but also ensure immediate and returnable safety of the entire state distribution when stochastically arriving at the new target state. 
A target state that satisfies this criterion of safe stochastic arrival is referred to as stochastic-arrivably safe. 
We first define a Single Step Safe Stochastic Arrival Operator $m$. 

\begin{definition}[Single-step Safe Stochastic Arrival Refinement Operator, $m$]
Given sets $S,R \subseteq X$, we define the single-step safe stochastic arrival refinement operator as:
\begin{align}
\label{eq:single_step_safe_stochastic_arrival}
    m^1(S, R) =& S \cup \{ x \in R | \exists z \in S, \action \in \actionSpace(x) \text{ s.t. } \\
    & \mathbb{P}(T(z, a) = x) > 0,\mathbb{P}(T(z, a) \in R) = 1\} 
\end{align}
\end{definition} 

The set $m^1(S, R)$ contains all the states in $R$ that the \agentspace can potentially reach from $S$ within a single step\footnote{We say potentially reach in a single step as the \agentspace may not reach the set target due to the stochastic transition function.}, without exiting $R$. 
To find the set of states in $R$ that can potentially be reached in a minimum of $n$ steps, without exiting $R$, we can recurse this this single step operator $m$.

\begin{definition}[Safe Stochastic Arrival Refinement Operator, $\safeStochArrivalOperator$] Given a control invariant set $S \subseteq X$ and $R = \safeStochReturnOperator(S, I(S))$, and single-step safe stochastic arrival refinement operator $m$, we define the safe stochastic arrival refinement operator $\safeStochArrivalOperator$ via recursion:
\begin{align} \label{eq:safeStochArrivalOperatorRecursion} 
    m^{n}(S, R) &= m^1(m^{n-1}(S, R), R) \nonumber \\
    \safeStochArrivalOperator(S, R) &= m^{|R|}(S, R) \nonumber
\end{align}
\end{definition}

Since $m^1$ is a restriction on the finite set $R$, the operator $\safeStochArrivalOperator$ is guaranteed to converge in at most $|R|$ steps.

In expansion $t+1$, to get the set of stochastic-arrivably safe states, we compute the set $\safeStochArrivalOperator(S_{t}, \safeStochReturnOperator(S_{t}, I(S_{t})))$, where $\safeStochReturnOperator(S_{t}, I(S_{t})))$ is the stochastic-returnably safe refinement of the immediate safe set $I(S_{t})$. 
This gives us the set of \textbf{completely} safe states at iteration $t+1$. 
We next show how to acquire this immediate safe set.


\subsection{Immediate Safe Expansion Operator}
Now that the safe stochastic refinement operators, $\safeStochReturnOperator$ and $\safeStochArrivalOperator$, are defined, we need to remove the oracle assumption and properly define the immediate safe expansion operator, $\bar{\immOperator}$. 
Recall that a state is immediately safe if its true safety value is less than a desired safety threshold, $\safetyFunction(\state) \leq \safetyThreshold$. 
As the true safety values, $\safetyFunction(\state)$, are unknown, they must be conservatively estimated with upper and lower confidence bounds $u_{t}(\state)\geq\safetyFunction_{t}(\state)\geq l_{t}(\state)$ that hold with high probability.
These tracked confidence bounds define two unique notions of immediate safety: \textbf{pessimistic} and \textbf{optimistic safety}, respectively; these break down into the following cases: 
\begin{itemize}
    \item If the upper bound is lower than the safety threshold, $u_{t}(\state) 
\leq \safetyThreshold$, then the true safety value at $\state$ will not exceed the safety threshold with a high probability. 
\item If the lower bound is higher than the safety threshold, $l_t(\state) \geq \safetyThreshold$, then the state is unsafe with high probability.
\item If $u_t(\state) > \safetyThreshold$ but  $l_t(\state) \leq \safetyThreshold$, there is some probability that additional information may reveal $\state$ to be safe. 
\end{itemize}
Thus, the lower bound $l_{t}(\state)$ informs the usefulness of continued exploration. Before we discuss the particular estimator used in this work to acquire $u_{t}(\state)$ and $l_{t}(\state)$, we will start by defining the immediate safe expansion operator assuming access to these bounds.



\begin{definition}[Single Step Immediate safe expansion operators $\immOperator^{p, 1}_{t}, \immOperator^{o, 1}_{t}$] Assuming a true safety function $\safetyFunction(\state)$ with given Lipschitz constant $L$, estimated upper and lower bounds such that $u_t(\state) \geq \safetyFunction_{t}(\state)\geq l_{t}(\state)$, and safety threshold $\safetyThreshold$, the
single step safe expansion operators $\immOperator^{p, 1}_{t}$ and $\immOperator^{o, 1}_{t}$ are defined as:
\begin{align}
    \immOperator^{p, 1}_{t} = \{ x \in \stateSpace | \exists z \in S~~ u_{t}(z) + Ld(x, z) + \epsilon \leq \safetyThreshold \}  \\
    \immOperator^{o, 1}_{t} = \{ x \in \stateSpace | \exists z \in S~~ l_{t}(z) + Ld(x, z) + \epsilon \leq \safetyThreshold \},
\end{align}
where $\epsilon \geq 0$ is an uncertainty term added to account for noise (recall $d: \stateSpace \times \stateSpace \rightarrow \mathbb{R}$ is the distance metric on $\stateSpace$). 
\end{definition}


These operators extend the confidence bounds around the safety value estimates of the known, current safe set by leveraging the Lipschitz assumption on how much the safety value can change across the state space to expand the pessimistic and optimistic safe sets. 
These single step expansions can be repeated until convergence to get the final set expansion operators. 

\begin{definition}\label{defn:imm_expansion}[Immediate Safe expansion operators $\bar{\immOperator}^{p}_{t}, \bar{\immOperator}^{o}_{t}$] 
Given a control invariant set $S \subseteq X$ and single step immediate safe expansion operators $\immOperator^{p, 1}_{t}, \immOperator^{o, 1}_{t}$, we define the safe expansion operators via recursion:
\begin{align}
    &\bar{\immOperator}^{p}_{t}(S) = 
     \immOperator^{p, |\stateSpace|}_{t}(S),\\
    &\bar{\immOperator}^{o}_{t}(S) = 
    \immOperator^{o, |\stateSpace|}_{t}(S),
\end{align}
where $\immOperator^{p, n}_{t}(S) := \immOperator^{p,1}_{t}(\immOperator^{p, n-1}_{t}(S))$ and $\immOperator^{o, n}_{t}(S) = \immOperator^{o,1}_{t}(\immOperator^{o, n-1}_{t}(S))$.
\end{definition}
Note that as $\stateSpace$ is a discrete finite state space, the recursion is indeed guaranteed to converge within $|\stateSpace|$ steps.

\subsection{Computing Confidence Bounds via Gaussian Processes} Having defined the expansion operator, we now introduce Gaussian Processes (GPs) to estimate and track the probabilistic bounds $u_t(\state), l_{t}(\state)$.
While our framework supports the use of any probabilistic estimation model, Gaussian Processes satisfy the requirements of a model that can readily be updated online with new data points and provide principled probabilistic estimates around the estimated values. 
In particular, we model the relationship between the state space and safety value using a GP Regression model. 

GPs are Bayesian non-parametric models that use Gaussian random variables to capture a distribution over the space of continuous functions. 
GPs are parameterized by a prior mean function, $\mu_{p}(\state)$ and a kernel function, $k(x, z)$: 
\begin{align}
    f(\state) = \mathrm{GP}(\mu_p(.), k(., .)) \\
    \mu_p(\state) = E[f(\state)] \\ 
    k(\state, z) = E[f(\state, z)]
\end{align}
Given training data, $\{ X, Y\}$ with training inputs $X \in \{ x_{0}, \dots, x_{n}\}$ and noisy training outputs: $Y \in \{ y_{0}, \dots, y_{n}\}$, which are modeled as $y_{i} = f(x_{i}) + \eta, \; \eta \sim \mathcal{N}(0, \sigma_{n}^{2})$, we can use the following equations to predict the posterior mean and variance at test points, $x^{*}$: 
\begin{align}
    f(x^{*}| x^{*}, X, Y) \sim \mathcal{N}(\mu(x^{*}), \sigma^{2}(x^{*})) \\
    \mu(x^{*}) = \mu_{p}(x^{*}) + K_{XX}^{-1}K(X, x^{*})\\
    \sigma(x^{*}) = k(x^{*}, x^{*}) - K_{XX}^{-1}K(X, x^{*}) \\
    K_{XX} = K(x^{*}, X)[K(X, X) + \sigma_{n}^{2} I]
    \label{eq:gp_regression}
\end{align}
Here $K(X, X) \in \mathbb{R}^{n \times n }$ is a covariance matrix constructed by computing the covariance between all training datapoints, where $K(X, X)[i, j] = k(x_{i}, x_{j})$. 
Similarly $K(X, x^{*}) = K(x^{*}, X)^{\top} \in \mathbb{R}^{n}$ are vectors computed using the covariance between each training datapoint in $X$ and the test point $x^{*}$: $K(X, x^{*})[j] = k(x_{j}, x^{*})$. 
In this paper we primarily use the Radial Basis Function (RBF) Kernel function: 
\begin{align}\label{eq:rbf_kernel}
    k(x, z) = \alpha \exp (-\frac{1}{2l^{2}} (x - z)^{\top}(x - z))
\end{align}
The scaling factor, $\alpha$, and lengthscale, $l$, are hyperparameters that specify the kernel function. 

We use these trained GPs to probabilistically estimate bounds around the safety values of states in the current safe set for the immediate safe expansion operators: $\bar{\immOperator}^{p}_{t}, \bar{\immOperator}^{o}_{t}$.
The GP is updated at every iteration, $t$, with new samples that are collected from the environment (this sampling procedure will be detailed in a future subsection). 
Following results from 
\cite{kernelizedmultiarmbandits}
the predicted distributions when using a scheduled beta value, $\beta_{t}$, gives us a confidence interval: $[l_{t}(x), u_{t}(x)]$ around the true safety value at a state. 
The upper bound, $u_{\iter}(\state) = \min(u_{\iter-1}, (\state) \mu_{\iter}(\state) + \beta_{\iter}\sigma_{\iter}(\state))$, of this interval provides a pessimistic notion of what the safety value could be while the lower bound, $l_{t}(\state) = \max(l_{\iter-1}(\state), \mu_{\iter}(\state) - \beta_{\iter}\sigma_{\iter}(\state))$, provides the optimistic notion. 
As shown in 
\cite{safeMDP}
$l_{\iter}(\state) \leq \safetyFunction(\state) \leq u_{\iter}(\state)$ with high probability when $\beta_{\iter}$ is chosen according to theorem 1 of 
\cite{safeMDP}. 
Note in Definition \ref{defn:imm_expansion} 
that the operators and the probabilistic bounds around safety are subscripted by the iteration $t$. 
This is as the probabilistic bounds and thus the immediate safe expansion operators change between iterations as they reflect our knowledge of the environment which improves as we collect more samples and further train our GP. 

While the above methodology provides guarantees around the probabilistically estimated safety value at the GP evaluated and immediately safe expanded states, we can introduce certain practical relaxations to ease these computations. 
While properly scheduling $\beta_{t}$ provides guarantees around the predicted confidence interval, the authors in 
\cite{safeMDP}
and 
\cite{safeIML}
show that the usage of a fixed $\beta_{t}$ is sufficient. 
Additionally the authors in \cite{safeIML} show that the computation of the immediate expansion operators  can be eased by directly using the GP to compute the bounds $u_{t}(x), l_{t}(x)$ at candidate states and comparing these values to the safety threshold, $u_{t}(x) \leq \safetyThreshold, l_{t}(x) \leq \safetyThreshold$ to dictate admittance to the pessimistic and optimistic safe sets respectively
\footnote{In the case of this relaxation, the choice of the GP hyperparameters warrants additional attention to ensure that unexplored states, distant from the existing safe set, are deemed unsafe, while sampling states at the boundary reduces enough uncertainty to grow the safe sets enough to learn about the unexplored environment}.


\subsection{\algname Discrete Case  Overview: Summary of Safe Set Expansion: }

Having defined all the operators, we can now summarize the complete safe set expansion process. 
At iteration $0$, we begin with a given starting safe set, $S_0$.
We set $S_0^p = S_0^o = S_0$.
We proceed iteratively as follows.
At iteration $t$, we start with $S_{t-1}^p$ and $S_{t-1}^o$.
The pessimistic and optimistic immediate expansion operators are then applied: $\bar{\immOperator}^{p}_{t}(S^{p}_{t-1})$ and $\bar{\immOperator}^{o}_{t}(S^{p}_{t-1})$. 
These set expansions capture the states that we infer to be pessimistically and optimistically immediately safe, given our assumptions on the environment and our current knowledge of the explored environment, as summarized by the updated GP. 
We then refine these sets to ensure safe stochastic return to the current safe set using the safe stochastic return operator: $\safeStochReturnOperator(S^{p}_{t-1}, \bar{\immOperator}^{p}_{t}(S^{p}_{t-1}))$ and $\safeStochReturnOperator(S^{p}_{t-1}, \bar{\immOperator}^{o}_{t}(S^{p}_{t-1}))$. 
This captures the sets of states that are pessimistically and optimistically immediately safe, and also allow the \agentspace to return to the current safe set $S^{p}_{t-1}$ without violating their notion of immediate safety. 
Finally, we ensure safe arrival from the current safe set using the safe stochastic arrival operator: $\safeStochArrivalOperator(S^{p}_{t-1}, \safeStochReturnOperator(S^{p}_{t-1}, \bar{\immOperator}^{p}_{t}(S^{p}_{t-1})))$ and $\safeStochArrivalOperator(S^{p}_{t-1}, \safeStochReturnOperator(S^{p}_{t-1}, \bar{\immOperator}^{o}_{t}(S^{p}_{t-1})))$. 
This gives us the updated pessimistic safe set and the updated optimistic safe set after the expansion at the current iteration: 
\begin{align}
\label{eq:discrete_full_safeset_expansion}
    S^{p}_{t} = \safeStochArrivalOperator(S^{p}_{t-1}, \safeStochReturnOperator(S^{p}_{t-1}, \bar{\immOperator}^{p}_{t}(S^{p}_{t-1}))) \\ 
    S^{o}_{t} = \safeStochArrivalOperator(S^{p}_{t-1}, \safeStochReturnOperator(S^{p}_{t-1}, \bar{\immOperator}^{o}_{t}(S^{p}_{t-1}))).
\end{align}
We can now use these sets to make decisions to safely sample the environment, update the GP, and recompute the safe sets for the next iteration.

\subsection{Guided Safe Goal-Driven Exploration}\label{sec:safe_goaldriven_exploration}

Given the expanded pessimistic and optimistic safe sets at iteration $t$, we now describe how to select the next state to sample in our safe, goal-directed exploration task. 
Let $x^*$ denote a given target state that we are interested in sampling, $\heuristicFunction(x, x^*)$ denote a heuristic function measuring the proximity between state $x$ and current target $x^*$, and $S^{p}_{t}, S^{o}_{t}$ denote the current expanded pessimistic and optimistic safe sets, respectively. 
Our sampling strategy builds on the method introduced in 
\cite{safeIML}. We begin by identifying sets of high-priority states: 
\begin{align}
    H_{t}(\alpha) = \{ \state \in S^{o}_{t} \backslash S^{p}_{t} | \heuristicFunction(\state, x^{*})  = \alpha \}
\end{align}
For the highest heuristic values, $\alpha$, this set contains states that are not yet verified to be safe to sample, but likely to yield high information about getting to the current target state $x^*$.
Since the states in $H_{t}(\alpha)$ are only optimistically safe, we cannot sample them directly. 
Instead, we aim to safely gather information about these states by sampling nearby points that are both uncertain and informative. 
To do this, we first define: 
\begin{align}
    U^{\delta}_{t} = \{\state \in \safeSet^{p}_{t}| u_{t}(\state) - l_{t}(\state) > \delta \}, 
\end{align}
the set of sufficiently uncertain states in $S^{p}_{t}$ where $u_{t}(x)$ and $l_{t}(x)$ are the upper and lower confidence bounds from the GP. 
From this, we construct the set of candidate expanders: 
\begin{align}
    G^{\delta}_{t}(\alpha) = \{x\in U^{\delta}_{t} | \exists z \in H_{t}(\alpha), l_{t}(\state) + Ld(x, z) \leq \safetyThreshold \}
\end{align}
The set $G^{\delta}_{t}(\alpha)$ includes pessimistically safe states that are both uncertain and within a Lipschitz-constrained distance to target states with heuristic value $\alpha$. 
Sampling states within $G^{\delta}_{t}(\alpha)$ helps inform better updates around the safety of states in $H_{t}(\alpha)$. 
We evaluate $G^{\delta}_{t}(\alpha)$ across different heuristic levels and select a state from the highest non-empty $\alpha$ set. 
We then plan a path to this state through the current pessimistic safe set $S^{p}_{t}$, ensuring that every transition stays within the pessimistic safe stochastic returnable set, denoted 
$\safeStochReturnOperator(S^{p}_{t-1}, \bar{\immOperator}^{p}_{t}(S^{p}_{t-1}))$, 
which guarantees both immediate safety and safe stochastic return. 
This guarantees task completion with rapid re-planning in the face of stochastic transitions. 
Upon reaching the selected state, we sample the safety function, update the GP model, and repeat the expansion planning process. 
Each added sample improves our probabilistic safety model, allowing us to better expand our safe sets. 
If the current desired target $x^*$ is found to be in the pessimistic safe set, it can be sampled directly.

For path planning, we use a simple deterministic transition model that selects the most probable next state for each transition. 
When the \agentspace detects that the next transition will exit 
the current set of safe stochastically returnable states, $\safeStochReturnOperator(\safeSet^{p}_{t-1}, \bar{\immOperator}^{p}_{t}(\safeSet^{p}_{t-1}))$, 
we employ rapid re-planning to ensure the \agentspace stays safe and can get to its current target. 
More sophisticated planners can be used, provided that they ensure that all planned transitions remain within 
$\safeStochReturnOperator(\safeSet^{p}_{t-1}, \bar{\immOperator}^{p}_{t}(\safeSet^{p}_{t-1}))$.


\section{Framework: Continuous Formulation}
In this section we consider the same general problem statement in the context of a continuous state space with a transition function, $T(x, a)$, characterized by a continuous probability distribution. 
To make the problem feasible, we restrict the problem to only consider Gaussian probability distributions for the transition function: $T(x, a) \sim \mathcal{N}(\mu(x, a), \Sigma(x, a))$. 
This Gaussian assumption is prevalent throughout literature when modeling stochastic dynamics for robotic~\cite{Thrun2005Probabilistic}. 
With this restriction, we can derive an expected transition function: $\bar{T}(x, a) = \mathbb{E}[T(x, a)] = \mu(x, a)$. 
This deterministic function can be used for planning, and allows us to consider each transition as a the expected, deterministic transition with the addition of zero mean Gaussian noise: $T(x, a) = \bar{T}(x, a) + \mathcal{N}(0, \Sigma(x, a))$.

\textbf{\algname Continuous Case: }Our overall solution for the continuous case takes a similar approach of having iteratively expanding safe sets. 
However, practically, tracking complex \textit{sets} in a continuous state space, without fine discretizations, can be prohibitively challenging.  
To mitigate this issue, we instead use indicator functions as functional descriptors of the safe sets. 
Specifically, we can define the safe set from a simple binary query (0: not belonging or 1: belonging to the safe set) of an underlying continuous safety function (derived from the GP posterior mean). 
We therefore must redefine the safe set expansion operators with these functional descriptions.

We maintain the immediately safe and safe stochastic arrival criteria of the discrete case. 
These ensure the safety of the \agentspace while in a state or trying to sample a state.  
The main change from the discrete case is that we relax the safe stochastic return criteria that ensures the \agentspace can safely continue its task objective. 
While this relaxation removes the guarantee of task completion from our solution, it still ensures that the \agentspace will remain instantaneously safe throughout its task. 

\subsection{Immediate Safe Expansion }
We consider a state to be immediately safe if its true safety value is below the safety threshold: $\safetyFunction(x) \leq \safetyThreshold$. 
As in the discrete case, we estimate the safety value at a state using a GP: $\hat{\safetyFunction}(x) = \mathrm{GP}(x) \sim \mathcal{N}(\mu(x), \sigma^{2}(x))$. 
However, unlike the discrete case, we do not maintain an explicit set of states that we can measure the distance from for Lipschitz continuity-based immediate safe expansion operators. 
Instead, we rely on the relaxation introduced in \cite{safeIML}, which allows us to directly employ the predicted distributions from the GP, $\mathcal{N}(\mu(x), \Sigma(x))$, to act as an immediate safe expansion indicator and determine if a state is in the expansion of the pessimistic or optimistic safe sets. 
If the predicted upper confidence bound on a state: $u_{t}(x) = \mu(x) + \beta_{t}\sigma(x) \leq \safetyThreshold$ the state is considered to be pessimistically immediately safe. 
If only the predicted lower confidence bound on a state is below the threshold, $l_{t}(x) = \mu(x) - \beta_{t}\sigma(x) \leq \safetyThreshold$, the state is considered to be optimistically immediately safe. 

\begin{definition}[Immediate Safe Expansion Indicators $\immIndicator^{p}_{t}, \immIndicator^{o}_{t}$]
Given a Gaussian probabilistic safety estimator $\hat{\safetyFunction}(\state) \sim \mathcal{N}(\mu(\state), \sigma^{2}(\state))$, the immediate safe expansion indicators are defined for state $\state$ as:
\begin{align}
\immIndicator^{\bullet}_{t}(\state) = \begin{cases}
    1 & \mu(\state) + \gamma_{\bullet}\beta_{t}\sigma(\state) \leq \safetyThreshold \\
    0 & \text{otherwise}
\end{cases}
\end{align}
where $\bullet \in \{p, o\}$, $\gamma_p = +1$, $\gamma_o = -1$, $\beta_{t}$ is an adjustable parameter, and $\safetyThreshold$ is the safety threshold. A value of $1$ indicates $\state$ belongs to the immediate safe set, and $0$ indicates otherwise.
\end{definition}

To ensure that the GP confidence intervals provide sufficient probabilistic bounds $\beta_{t}$ is scheduled as in \cite{safeIML}. 
However, in practice, safety can be maintained by setting a fixed $\beta$ of sufficient magnitude, 
demonstrated in \cite{safeIML}.
This relaxation is made feasible because a GP is a non-parametric method that relies on distance-based similarity between a new state and states in its training distribution to extrapolate its safety value. 
With the right choice of kernel hyperparameters, such as the lengthscale of an RBF kernel, this can allow new samples to naturally extend a safe set in a manner that approximates the Lipschitz continuity and distance-based safe expansion for immediate safety. 

\subsection{Safe Stochastic Arrival}
The criteria of safe stochastic arrival ensures that when planning to sample some state, $x$, not only is $x$ immediately safe, but the entire distribution that the \agentspace can arrive at after a transition to $x$ is safe with a high probability. 
While one approach to ensure this in a continuous state space is through evaluating samples of the distribution after every transition, this tends to be quite computationally demanding. 
Instead, by relying on Lipschitz continuity assumptions of the safety function and the Gaussian assumption on the stochasticity of the transition function, we can analytically compute the safety value bounds around a full arrival distribution. 

In our problem statement, the state after a transition is a random variable $\mathbf{x}_{t+1} = T(x_{t}, a)$. 
Particularly $\mathbf{x}_{t+1}$ is a random variable with a Gaussian Distribution: $\mathbf{x}_{t+1} \sim \mathcal{N}(\mu(x, a), \Sigma(x, a))$. 
When predicting safety, the GP is constructed with the assumption that all the safety values are distributed according to a multivariate Gaussian assumption. 
In standard GP regression, we solve for the distribution of the safety value at an individual test state, $x'$: $\mathrm{GP}(x') \sim p(\safetyFunction(x')) \sim \mathcal{N}(\mu(x'), \sigma^{2}(x')) $. 
The GP Regression equations in eq \eqref{eq:gp_regression} solve for the marginal distribution of the safety value a particular desired test point, $x'$: $p(\safetyFunction(x')) = p(\safetyFunction(x') | x', X, \safetyFunction(X)))$, where $\{X, \safetyFunction(X)\}$ is the GP's training data. 
When performing the same prediction for a random variable input $\mathbf{\state'}$: 
\begin{align}
    p(\safetyFunction(\mathbf{\state'})) = \int p(\safetyFunction(\mathbf{\state'})) | \mathbf{\state'}, X, \safetyFunction(X)) p(\mathbf{\state'}) d \mathbf{\state'}
\label{eq:intractable_integral}
\end{align} 
we integrate out the random variable $\mathbf{\state'}$. 
Solving the integral in equation~\eqref{eq:intractable_integral} is intractable. 
Inspired by approaches in \cite{nonparametricvideo,  PILCO}, we approximate $p(\safetyFunction(\mathbf{\state'}))$ as a Gaussian, with distribution $\mathcal{N}(\mu_{\mathrm{stoch}}(\mathbf{x'}), \sigma^{2}_{\mathrm{stoch}}(\mathbf{x'}))$, and use moment matching to derive an analytical solution to equation~\eqref{eq:intractable_integral}. 
To solve for this analytical solution, we rely on the assumptions that the input random variable is distributed by a Gaussian distribution and that the underlying safety function, $\safetyFunction$, can be represented by a GP using a Radial Basis Function (RBF) Kernel as specified in equation~\ref{eq:rbf_kernel}. 

The analytical solution of the GP estimated safety value at the random variable state $\mathbf{\state}' \sim \mathcal{N}(\mu', \Sigma')$ is specified by the distribution $p(\safetyFunction(\mathbf{\state'})) \sim \mathcal{N}(\mu_{\mathrm{stoch}}(\mu', \Sigma'), \sigma^{2}_{\mathrm{stoch}}(\mu', \Sigma'))$ where: 
\begin{align}\label{eq:mu_stoch}
    \mu_{\mathrm{stoch}}(\mu', \Sigma') &= d^{T} B \\
    d[i] &= \alpha^{2} (\Sigma' \Lambda^{-1} + I)^{-\frac{1}{2}} e^{-\frac{1}{2} v_{i}^{T} (\Sigma' + \Lambda)^{-1} v_{i}}\nonumber \\
    B &= [K + \sigma^{2}_{n}I]^{-1} \safetyFunction(X) \nonumber \\
    v_{i} &= X[i] - \mu' \nonumber
\end{align}
$\mu_{\mathrm{stoch}}(\mu', \Sigma') \in \mathbb{R}, d \in \mathbb{R}^{n \times 1}, B \in \mathbb{R}^{n \times 1}$ where $n$ is the number of datapoints in the training dataset. 
\begin{align}\label{eq:sigma_stoch}
    \sigma^{2}_{\mathrm{stoch}}(\mu', \Sigma') &= \alpha^{2} - tr((K + \sigma^{2}_{n}I)^{-1}Q) \nonumber \\
    &\;\;\;\;  + B^{T}QB - \mu_{\mathrm{stoch}}(\mu', \Sigma') \nonumber \\
    Q[i, j] &= \frac{k(X[i], \mu') k(X[j], \mu')}{\sqrt{|R|}} e^{\frac{1}{2} z_{i,j}^{T} R^{-1} \Sigma z_{i, j}} \nonumber \\ 
    R &= \Sigma'(\Lambda^{-1} + \Lambda^{-1}) + I \nonumber \\ 
    z_{i, j} &= \Lambda^{-1}v_{i} + \Lambda^{-1}v_{j} \nonumber
\end{align}
Here $\sigma^{2}_{\mathrm{stoch}}(\mu', \Sigma')\in \mathbb{R}, Q \in \mathbb{R}^{n \times n}$.
$\Lambda = l \cdot I $ is a diagonal matrix where $l$ is the RBF Kernel lengthscale, $\alpha$ is another RBF kernel parameter and $k$ is the RBF kernel function. 
$X, q(X)$ is the data used to train the GP with $X[i]$ referring to the $i^{\text{th}}$ input in the training dataset. 
$K = k(X, X) \in \mathbb{R}^{n \times n}$ is the matrix of the kernel values between all training data points, $K[i, j] = k(X[i], X[j])$.
$\sigma_{n}$ is the noise variance of the GP. 

We use equations~\eqref{eq:mu_stoch},~\eqref{eq:sigma_stoch} to directly predict the safety value of an arrival distribution create an indicator function to directly determine safe stochastic arrival. 
Given a random variable state, $\mathbf{\state}_{t} = T(\state, a)$, with distribution $p(\mathbf{\state}_{t}) \sim \mathcal{N}(\mu', \Sigma')$, where $\mu'=\mu(x, a)$ and $\Sigma'=\Sigma(x,a)$, we can use the GP to construct upper and lower confidence bounds, $u_{t,\mathrm{stoch}}(\mu', \Sigma') = \mu_{t, \mathrm{stoch}}(\mu', \Sigma') + \beta_{t}\sigma_{t, \mathrm{stoch}}(\mu', \Sigma')$ and $l_{t,\mathrm{stoch}}(\mu', \Sigma') = \mu_{t, \mathrm{stoch}}(\mu', \Sigma') - \beta_{t}\sigma_{t, \mathrm{stoch}}(\mu', \Sigma')$ respectively. 
Akin to the immediate safe expansion  indicators, if the upper bound is lower than the safety threshold, $u_{t, \mathrm{stoch}}(\mu', \Sigma') \leq \safetyThreshold$ we consider the arrival distribution to be pessimistically safe. 
Similarly, if the lower bound is under the safety threshold, $l_{t, \mathrm{stoch}}(\mu', \Sigma') \leq \safetyThreshold$, the distribution is considered optimistically safe. 
 This effectively yields a GP-based indicator function that classifies the safety of the arrival distribution.

However, our goal is to directly assess whether arriving at a specific state $x_{t}$ is safe. 
To do so, we again consider the transition function $T(x, a) \sim \mathcal{N}(\mu(x,a), \Sigma(x, a))$. 
Recall that we can decompose the transition dynamics into a deterministic component, $\bar{T}(x, a) = \mathbb{E}[T(x, a)] = \mu(x,a)$ and additive zero-mean Gaussian noise $\mathcal{N}(0,\Sigma(x,a))$. 
Since planning follows the deterministic, expected model we can conservatively identify the transition with the greatest variance when transitioning to the state of interest, $x$, and use its associated variance, 
$\Sigma_{\max}(\state) =\max \{\Sigma(z, a) \; | \; \forall z \; s.t. \bar{T}(z, a) = \state \}$,
to construct the worst case arrival distribution: $\mathcal{N}(x, \Sigma_{\max}(\state))$. 
We can then use the GP, as shown above, to compute the safety bounds of this distribution and create an indicator function that determines whether $x_{t}$ satisfies Safe Stochastic Arrival for either pessimistic or optimistic safety criteria. 
Notably, since this condition also ensures immediate safety of the state, we can directly use this indicator as the sole determinant of the safe set. 

\begin{definition}[Safe Stochastic Arrival Indicators $\safeStochArrivalIndicator^{p}_{t}, \safeStochArrivalIndicator^{o}_{t}$]
Given a Gaussian Process safety estimator $\hat{\safetyFunction}(\state) = \mathrm{GP}(\state) \sim \mathcal{N}(\mu(\state), \sigma^{2}(\state))$, and an expected arrival state, $\state$, the safe stochastic arrival indicator functions are:
\begin{align}
\safeStochArrivalIndicator^{\bullet}_{t}(\state) = \begin{cases}
        1 & \mu_{\mathrm{stoch}}(\state, \Sigma_{\max}) + \gamma_{\bullet}\beta_{t}\sigma_{\mathrm{stoch}}(\state, \Sigma_{\max}) \leq \safetyThreshold \\
        0 & \text{otherwise}
    \end{cases}
\end{align}
where $\bullet \in \{p, o\}$, $\gamma_p = +1$, $\gamma_o = -1$, $\beta_{t}$ is an adjustable parameter, $\safetyThreshold$ is the safety threshold, $\mu_{\mathrm{stoch}}$ is defined in Eq.~\ref{eq:mu_stoch}, $\sigma_{\mathrm{stoch}}$ is defined in Eq.~\ref{eq:sigma_stoch}, and $\Sigma_{\max}(\state) =\max \{\Sigma(z, a) \; | \; \forall z \; s.t. \bar{T}(z, a) = \state \}$ represents the greatest stochasticity when transitioning to state $\state$. 
A value of $1$ indicates $\state$ belongs to the safe stochastic arrival set, and $0$ indicates otherwise.
\end{definition}

\subsection{\algname Continuous Case: Summary of Expansions}

We use the safe stochastic arrival 
indicators, $\safeStochArrivalIndicator^{p}_{t}, \safeStochArrivalIndicator^{o}_{t}$,
to expand safe sets in the continuous state space. 
As in the discrete case, we assume the robot begins in a control-invariant safe set, with samples available to train an initial GP model. 
A sampling-based planner \cite{karaman2011samplingbasedalgorithmsoptimalmotion}, such as Rapidly exploring Random Trees (RRT) \cite{LaValle1998RapidlyexploringRT} or Probabilistic Roadmaps (PRMs) \cite{PRMs}, constructs a graph of discrete states rooted at the current state, aiming to reach the goal. 
Planning and graph construction are performed using the expected deterministic transition model. 


We initialize both the pessimistic and optimistic safe sets, $\safeSet^{p}_{t}$ and $\safeSet^{o}_{t}$, with the current robot state. These sets are then expanded outward along the constructed graph. 
A state $\state$ is added to $\safeSet^{p}_{t}$ if there exists a parent state $z \in \safeSet^{p}_{t}$ and an action $a$ such that the expected transition $\bar{T}(z, a)$ reaches $\state$ and the stochastic arrival distribution satisfies the pessimistic safety criterion:
\begin{align}\label{eq:continuous_pessimistic_safe_set_expansion}
    \safeSet^{p}_{t} = \{\state_{t} | \exists z \in \safeSet^{p}_{t}, s.t. \; \\ \nonumber
    \Bar{T}(z, a) = \state_{t}, u_{\mathrm{stoch}}(\mu(z, a), \Sigma(z, a)) \leq \safetyThreshold \}
\end{align}
This formulation can be adapted to enforce stochastic continuation if required. The optimistic safe set $\safeSet^{o}{t}$ is constructed analogously, using the other probabilistic bound for the optimistic safety criteria, and satisfies the inclusion $\safeSet^{p}{t} \subset \safeSet^{o}_{t}$ by definition.

The expanded safe sets, $\safeSet^{p}_{t}, \safeSet^{o}_{t}$, are used to guide goal-directed exploration. 
We select a target state from the frontier and plan a path to it using $\bar{T}$. 
To find the target set we follow the same safe goal-driven exploration procedure described in sectiion~\ref{sec:safe_goaldriven_exploration}. 
Due to stochastic dynamics, before every planned action, we verify that the resulting state remains within the safe set under the pessimistic safe stochastic arrival criteria, using $\safeStochArrivalIndicator^{p}_{t}$. 
If the system deviates from the safe set due to stochasticity, we re-plan or sample at the current state and repeat the safe set expansion and sampling process.

\subsection{Safety Adjustments for a Continuous State Space }
Although we consider discrete actions, the underlying state space is continuous, which requires ensuring safety not just at discrete states but also along the trajectories connecting them. To address this, we leverage the Lipschitz continuity of the safety function $\safetyFunction$ to guarantee path-wise safety.

Since $\safetyFunction$ is Lipschitz continuous with constant $L$, we have: $|\safetyFunction(x) - \safetyFunction(z)| \leq L d(x,z)$.
This bounds the maximum change in safety between two nearby states.
Now consider the scenario when the \agentspace is moving between two states, $x_{t}$ and $x_{t+1}$ where the safety value is at the safety threshold: $\safetyFunction(x_{t}) = \safetyFunction(x_{t+1}) = \safetyThreshold$. 
We need to ensure that the \agentspace does not violate the safety threshold along its path.
Here the worst case scenario is where the safety function increases and then decreases at the maximum rate $L$. 
The peak deviation from the safety threshold $\safetyThreshold$ along a transition is at most $L \cdot d(\state_t, \state_{t+1})/2$.
To account for this, we can conservatively adjust the safety threshold used during planning to: $\safetyThreshold^{'} = \safetyThreshold - L d_{\max}/2$. 
Here $d_{\max} = \max_{a} d(\state, T(\state, \action))$ denotes the maximum step size induced by any action.
Too large a $d_{\max}$ can increase conservatism by drastically reducing the permissible safety threshold. 
We can regulate $d_{\max}$ by limiting the action resolution or step size in the sampling-based planner used to build the state graph.
This ensures that if two connected states in the graph both satisfy the adjusted threshold $\safetyThreshold'$, then the entire path segment between them is guaranteed to remain safe.


\subsubsection{Probabilistic Relaxation}
We introduce a probabilistic relaxation to balance safety and goal-directed exploration according to user preferences. 
In some cases, the pessimistic safety criterion can be overly conservative, limiting exploration and preventing goal reachability. 
To mitigate this, we scale down $\beta$.  
This reduces the number of standard deviations considered when evaluating safety, allowing more states to be classified as safe under uncertainty.
This parameter can be set dynamically in instances when the \agentspace gets stuck or no new expansion states are found to balance the tradeoff between task completion and safety. 
It can be reported back to a user allowing the user to balance the tradeoff between goal completion and safety before the \agentspace moves. 
Alternatively, it can be autonomously reduced up to some minimum threshold, $\beta = \beta_{\text{scaling}} \cdot \beta$ while $\beta' \geq \beta_{\mathrm{min}}$, with $\beta_{\text{scaling}} < 1.0$, $\beta_{\text{scaling}} = 0.99$.

\section{Framework: Object-Centric Formulation}
\label{sec:object_case}
In this section we present an application-driven instance of the general safe goal-driven exploration problem, and outline how to build upon our methods from the discrete and continuous cases to apply them to these scenarios. 
Enabling robots to safely operate around objects in homes and other unstructured environments is a key motivation behind the methodology introduced in this paper. 
To illustrate this, we examine a tabletop robotics scenario to serve as a concrete example whose concepts can be generalized to other domains. 
Consider a tabletop with opaque bottles and containers. 
The objects are made to appear visually identical, with similar geometries, but have differing hidden physical properties such as friction coefficients, mass or center of mass. 
These differences affect how a robot interacts with these objects. 
The robot is tasked with safely reaching a desired goal location on the tabletop, which requires the robot to physically interact with these objects. 
However, while moving across the table, the robot must avoid creating unsafe scenarios through safety violations, which we define as tipping any object beyond a pre-specified angle. 
Such tipping could cause the container's contents to spill or even fall over and break. 
This scenario is designed to mimic the unstructured environments that a robot may encounter when interacting with unknown, constantly changing objects in unstructured environments like a household shelf or pantry. 
Importantly, the robot does not \textbf{1)} know the properties of the objects a priori \textbf{2)} have prior knowledge of its own interaction dynamics with each object. 



\subsection{Single Object Problem Formulation:}
To formally state the problem, we first consider the case of a single unknown object. 
We consider a robot that moves quasi-statically and has an end-effector capable of non-prehensile interactions with the object. 
Since the object's properties are unknown, we model the robot-object interactions using a stochastic transition model, $T$. 
The system state is: $[\state_{r}, \state_{o}]$, where $\state_{r}$ is the robot end-effector state and $\state_{o}$ is the object state. 
The robot's objective is to reach a desired goal region, guided by a heuristic function $\heuristicFunction$, while maintaining safety. 
The safety function is $\safetyFunction(\state_r, \state_o)$ which returns the object's tipping angle. 
The robot remains safe if $\safetyFunction(\state_r, \state_o) \leq \safetyThreshold$ where $\safetyThreshold$ is the maximum allowable tipping angle. 
We assume $\safetyFunction$, while within the safe range of $q(\state_{r}, \state_{o} \leq \safetyThreshold)$, is Lipschitz continuous with constant $L$ under a given positive definite distance function $d$ on the joint state space: 
$\safetyFunction([\state_{r}, \state_{o}]) - \safetyFunction([\state_{r}, \state_{o}]) \leq L d([\state_{r, 0}, \state_{o, 0}], [\state_{r, 1}, \state_{o, 1}])$. 

\subsection{Multi-Object Problem Formulation: }
To extend to $n$ objects, the environment state is defined as $[\state_{r}, \state_{o,1}, \dots, \state_{o, n}]$. 
The safety function, $\safetyFunction$, now captures the worst-case scenario: meaning the robot must ensure that no object exceeds the tipping angle set by the safety threshold, $\safetyThreshold$. 
To relax the problem we make several simplifying assumptions. 
We do not consider the effects on safety of object-object interactions, assuming them to be minimal. 
We assume a uniform environment where objection motion is solely determined by robot-object interactions. 
Furthermore we assume that the robot end-effector is capable of exerting enough force to readily move any objects in the environment. 

With these assumptions, the tabletop safe object problem maps directly to the general safe goal-driven exploration problem. 
The \agent's state is defined by the robot state $\state_r$, while the underlying safety function, $\safetyFunction$, is dynamically conditioned on the (possibly changing) states of the surrounding objects $[\state_{o,1}, \dots, \state_{o, n}]$. 
Given a fixed object configuration, we can either apply the continuous-case solution or discretize the state space to employ the discrete-case solution to perform safe set expansion for the safe goal driven exploration task. 
In both cases, the safety function fully captures the relevant effects of the robot-object interactions, allowing the robot to be treated as though it is effectively moving through "free space".

Although this theoretical framing renders the tabletop problem structurally identical to the general problem statement—and therefore compatible with the solutions described above—the practical use of GPs to implicitly model safety in the presence of unknown physical interactions introduces two key challenges. 
\textbf{(1)} Object positions change continuously as a result of the robot’s interactions, making it difficult for non-parametric models such as GPs to effectively leverage past interaction data. 
\textbf{(2)} The challenge intensifies with multiple moving objects, since safety depends on the full environment state, which consists of the robot state along with the state of each object. 
This not only increases the complexity of set-based planning, but also compounds the difficulty for the GP to generalize from prior interactions. 
The remainder of this section describes our strategies for overcoming these challenges.

\subsubsection{\algnamenospace: Single Moving Object} We first address the challenge of applying our above solutions to environments with a single moving object. 
GPs are non-parametric models that rely on kernel-based similarity metrics between new inputs and previously sampled data points. 
If we directly apply a kernel over the joint robot-object state, $k([\state_{r}, \state_{o}], [z_{r}, z_{o}])$, changes in the object's state, caused by robot interactions, will cause future inputs to drift far from the GPs support set. 
This reduces the GP's ability to accurately predict safety values for new states. 

The key insight is that safety often depends not on the absolute states of the robot and object, but on their relative configuration. We therefore adopt an object-centric representation  when modeling safety with GPs, using the relative state of the robot in the object's coordinate frame, $\state_{ro}$~\cite{shinde2023objectcentricrepresentationsinteractiveonline}. 
Under the assumption of a uniform environment, this representation ensures that past data remains valid even as the object moves, as relative states preserve the underlying interaction patterns, enabling the GP to make consistent predictions. 
Importantly, while the GP predictions and the safe set expansions occur in relative coordinates, the task and thus the heuristic function still operate on the absolute state of the robot. 

\subsubsection{\algnamenospace: Multiple Moving Objects} For environments with $n$ moving objects, we decompose the problem. 
Safety is violated if any object becomes unsafe (e.g., tips over beyond a set threshold), so the global safety function is the maximum over the individual safety functions of each object. 
Assuming safety depends solely on the robot-object interactions, each object can be treated as an independent instance of the continuous or discrete safe exploration problem. 
We model each object's safety independently with a GP over its own relative object-centric state space, compute the safe set for each object
: $[\{\safeSet^{p}_{t, 1}, \safeSet^{o}_{t, 1}\}, \dots \{ \safeSet^{p}_{t, n}, \safeSet^{o}_{t, n}\}]$, 
and then take the intersection of all safe sets to obtain the overall safe set
: $\safeSet^{p}_{t} = \cap_{i=1}^{n}\safeSet^{p}_{t, i}, \safeSet^{o}_{t} = \cap_{i=1}^{n}\safeSet^{o}_{t, i}$
. 
The overall safe set and thus this intersection is taken in the absolute state space of the robot.
This intersection guarantees that all objects remain safe while the robot is contained within the safe set. 
Once the intersection is computed, we can apply the same safe goal driven exploration procedures as the discrete and continuous case solutions (as outlined in section~\ref{sec:safe_goaldriven_exploration}) to select the next sampling point, iterating until the desired goal is reached. 
This decomposition also ensures consistent monotonic safe set expansions for each individual decomposed problem; however, the intersection may result in the safe set shrinking between iterations due to the object's physical movement. 
Also note that while intersection preserves safety guarantees, it compromises safe stochastic return guarantees, that is the ability to ensure a safe return path under uncertainty is lost as the intersection may prune states necessary to ensure safe return as objects move through the environment. 

\subsection{Practical Considerations }
In practice, the continuous case is generally better suited for fine-grained interactions
. 
When addressing multiple-object problems in the continuous case, we employ a single discrete state graph in the absolute space and transform it into each object’s relative space, which simplifies the computation of intersections. 
Rather than explicitly solving separate set-expansion problems for each object and intersecting the results, we instead modify the probabilistic safety predictor in the original method. 
In this approach, a separate GP is trained for each object using relative-state inputs, and at prediction time the results are aggregated by selecting the worst-case probabilistic bound across all objects. 
This yields the same outcome as intersecting the independently computed safe sets but avoids the computational cost of solving multiple problems. 
However, as objects move and the safety map evolves, the set of safe states in absolute coordinates can shrink. 
Consequently, if this aggregation-based approach is used instead of solving independent problems, the safe set must be re-expanded from the current object configuration—making the method more naturally suited to the continuous-case solution.

\section{Experiments and Results}
In this section we present the experiments conducted, both in simulation and hardware, to validate all variants of \algnamenospace. 
Our simulation experiments are conducted across $2$ separate environments. 
We begin by describing the experiments conducted in the Ground Robot Environment, used to validate the discrete and continuous case formulations of our method. 
We then describe the experiments in the Safe Object Environment, used to validate the object case of our method. 
Finally we detail the hardware setup and corresponding real world experiments.

\subsection{Ground Robot Environment Experiments}
\subsubsection{Ground Robot Environment} 
\begin{figure*}[t]
    \centering

    \includegraphics[width=\linewidth]{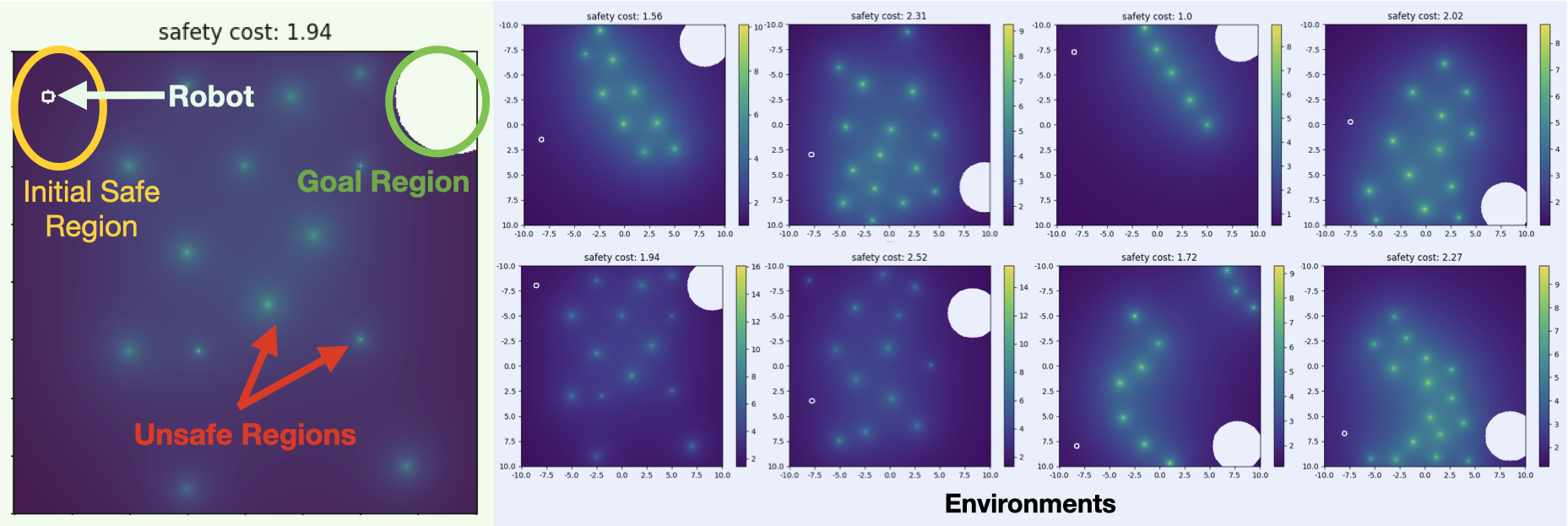}

    \caption{\textbf{Ground Robot Environments: }
    All $8$ environments used for the Ground Robot Experiments. 
    In every environment, we show the robot randomly initialized within its initial safe region. 
    The color represents the ground truth safety value (with higher values indicating more unsafe conditions), with a safety threshold of $4$ across all environments.}
    \label{fig:ground_envs}
\end{figure*}

Visualized in Fig.~\ref{fig:ground_envs}, our first environment involves a ground robot navigating a bounded 2D grid: $x,y \in [-10, 10]$, inspired by the Mars Rover environment in \cite{safeIML}. 
The \agent's state, $\state_r$, is its $(x,y)$ position. 
The \agentspace can take discrete actions (up, down, left, right), with a specified step size $s$. 
Each transition is subject to random noise and actions that would exit the grid boundaries are not allowed. 
The specifics of the transition function varies for each experiment and are detailed in later subsections. 

The environment contains unsafe regions (e.g., craters, quicksand, landmines), each defined by a center state $\state_o$. 
The safety of each state in the environment is captured by a safety function, $\safetyFunction$.
Intuitively, this function may represent the crater's slope, propensity to get stuck, or be a function of landmine proximity. 
We assume access to sensor readings of this function. 
A state is unsafe if its safety value exceeds a safety threshold of $4$, corresponding to irrecoverable conditions (e.g., tipping, getting stuck, landmine detonations). 
The ground truth safety function for the ground robot environment is implemented as: 
\begin{align}
    \safetyFunction(\state_{r}) =
    \sum_{i=1}^{n} 
    \mathbf{1}\!\left( \|x_{r} - x_{o, i}\|_{2} \le d_i \right) 
    \frac{1}{\|x_{r} - x_{o, i}\|_{2}^{2} + \frac{1}{c_{i}}}
\end{align}
where $n$ is the number of unsafe regions, $c_i$ is the maximum cost at the unsafe region center ($x_o$), and $d_i$ is a cutoff radius beyond which the $i^{\text{th}]}$ unsafe region has no effect. 
The \agent's objective is to reach a goal region centered at $x_g$ with radius $r_g$ without violating the safety threshold. 
This environment is implemented in python with a Gym-style interface with the addition of the safety function. 
This environment is used to test both the discrete and continuous case methods. 

\subsubsection{Discrete Case Experiments: }\label{sec:discrete_case_experiments}
We validate \algnamenospace's discrete case method using the ground robot environment. 
The environment is discretized into a 2D grid state space using a step size of $s=0.25$. 
The robot state is specified with $x,y$ grid coordinates. 
The robot can take up, down, left and right discrete actions with a step size of $s=0.25$. 
These actions move the robot according to a stochastic transition function, $T$, characterized by a discrete probability distribution for this set of experiments. 
For the discrete action space (up, down, right, left) the \agentspace moves to the intended states, $y \mathrel{+}= s, y \mathrel{-}= s, x \mathrel{+}= s, x \mathrel{-}= s$, with probability $p$. 
With probability $(1-p)/2$ the \agentspace transitions to one of the two adjacent diagonals: $(x\mathrel{+}=s, y\mathrel{+}=s)$ or $(x\mathrel{-}=s, y\mathrel{+}=s)$ for up, $(x\mathrel{+}=s, y\mathrel{+}=s)$ or $(x\mathrel{+}=s, y\mathrel{-}=s)$ for right etc. 
The \agentspace planner simply uses the intended state with probability $p$ and considers a deterministic transition. 
The \agentspace uses a naive heuristic function based on Euclidean distance to the goal center, $x_g$. 

We conducted $80$ random trials across $8$ challenging environments, with distinct placements of the unsafe regions. 
These environments are shown in Fig. ~\ref{fig:ground_envs}. 
In each experiment the robot is randomly initialized within a start region, and must get within radius $r_g$ of the goal region, $x_g$, without exceeding a safety threshold of $\safetyThreshold = 4$. 
We test with extremely high stochasticity with $p$ values set to $0.35$. 
At this value the intended transition remains the most likely by a small margin.

We compare \algname against a baseline method from \cite{safeIML} that does not account for stochasticity. 
The baseline is modified for fairness, allowing it to check GP estimated safety at its next intended state, and re-plan if the next state is detected to be unsafe. 
Runs share the same random seed across methods.

We evaluate each algorithm using metrics of success rate (higher better), violation rate (lower better) and number of states explored (higher better). 
A success here means that the robot has reached the goal region without exceeding the safety threshold. 
Safety violation rate captures the number of safety violations over trials, and number of states explored, captures how long the robot was able to continue exploring before either reaching the goal or having a safety violation. 
The discrete case results, summarized in Table~\ref{tab:discrete_case_table}, demonstrate the robustness of our method even under such noise. All trials from the results in this table were run with the same GP and algorithm configuration. 
The results highlight how, by properly considering stochasticity, \algname is consistently able to reach the goal without any safety violations. 
Meanwhile, the baseline frequently fails and violates safety. 
An example trajectory of the discrete case of both the baseline and our method is shown in Fig.~\ref{fig:ground_traj}.

\begin{table}[h]
\centering
\caption{\textbf{Discrete Case Results: }  
From Discrete Case Ground Robot experiments in Section~\ref{sec:discrete_case_experiments}. 
}
\label{tab:discrete_case_table}
    \begin{tabular}{lcccc}
    \hline 
    \textbf{Method} & \textbf{Success \%} & \textbf{Violation \%} & \textbf{Avg. States Explored} \\
    \hline 
    Baseline & 35\% & 65\% & 563.91 \\
    \textbf{\algname} & 100\% & 0 \% & 1747,33 \\
    \hline 
    \end{tabular}
\end{table}

    
    
    
    
    
    

\begin{figure*}
    \centering
    \begin{subfigure}[t]{0.95\linewidth}
      \centering
      \resizebox{\linewidth}{!}{\includegraphics{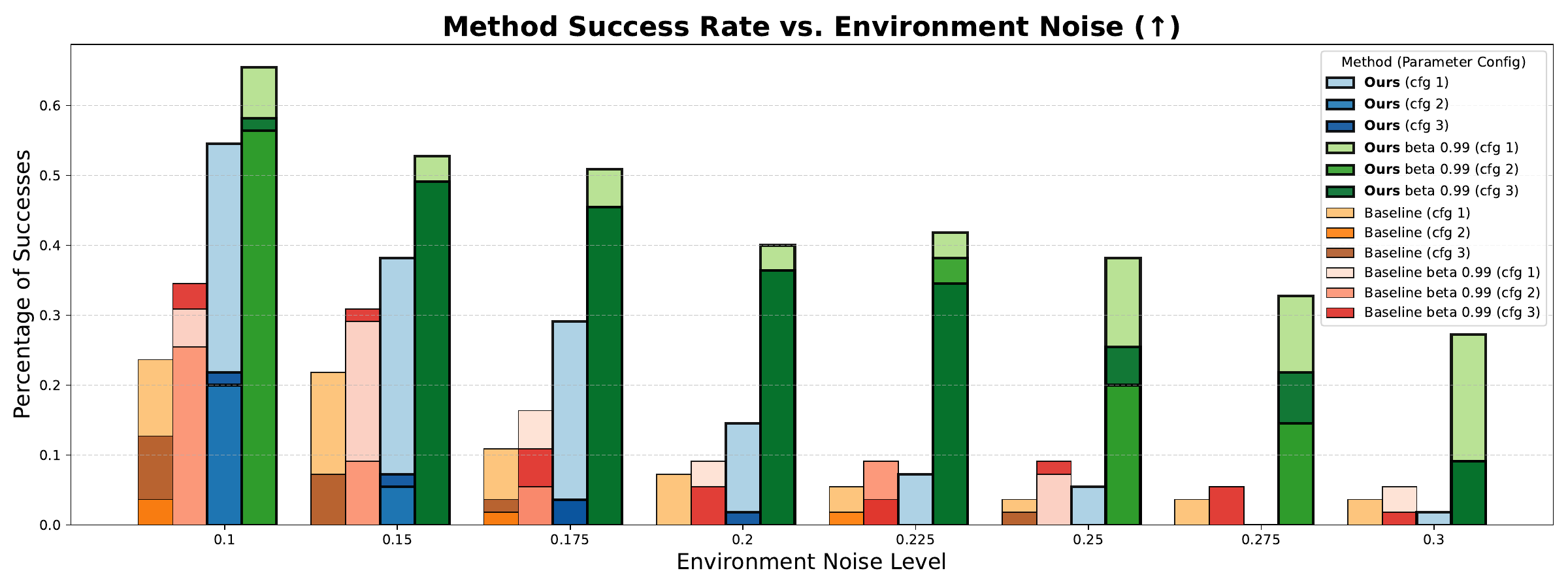}}
    \end{subfigure}%

    \vspace{0.001em}

    \begin{subfigure}[t]{0.95\linewidth}
      \centering
      \resizebox{\linewidth}{!}{\includegraphics{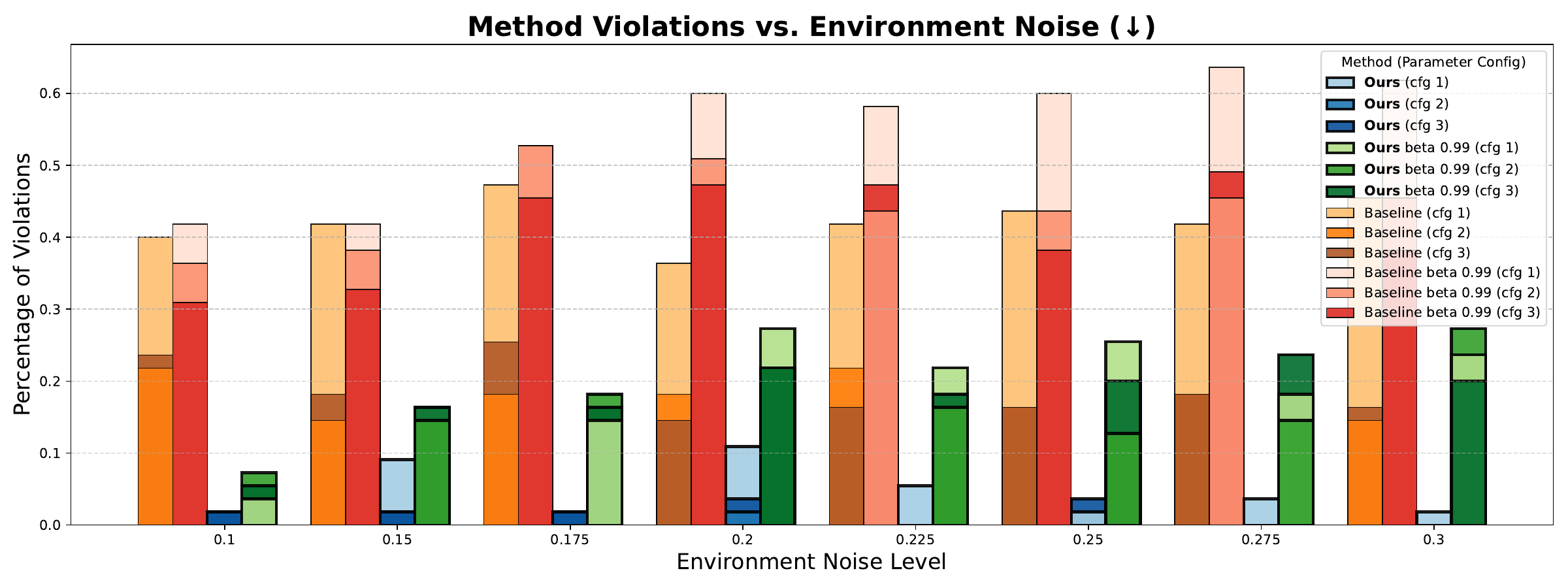}}
    \end{subfigure}%

    \vspace{0.001em}

    \begin{subfigure}[t]{0.95\linewidth}
      \centering
      \resizebox{\linewidth}{!}{\includegraphics{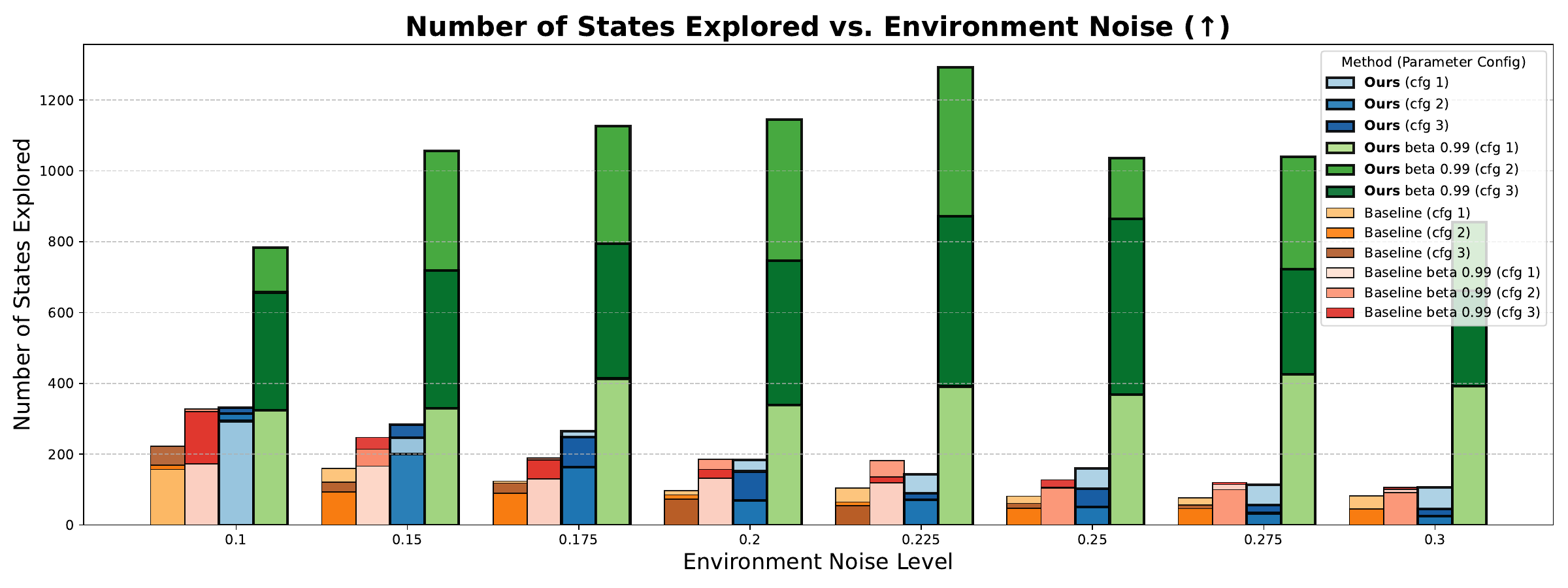}}
    \end{subfigure}%
    \vspace{-0.5em}
    \caption{\textbf{Continuous Case Growing Stochasticity Ground Robot Experiment Results:} detailed in Section~\ref{sec:continuous_case_experiments}. 
    The plots compare the success rate, violation rate and average states explored across increasing levels of environment noise in the Ground Robot Environments. 
    Each individual bar captures the results across $40$ randomized trials over $8$ environments. 
    Performance is shown for our method (blue/green) and the baseline (orange/red), using both normal and $\beta$ scaling variants. 
    Each shade of color represents one of $3$ GP configurations. 
    As environment stochasticity increases, our method with $\beta$ scaling maintains substantially higher success rates and lower safety violation rates in comparison to either baseline. 
    this is further supported by the larger number of states explored before termination. 
    $\beta$ scaling enables us to effectively manage the safety-performance tradeoff, while our base method prioritizes safety and becomes too conservative and loses performance as noise grows (though still outperforming the baselines) . 
    Showing performance across different GP configurations demonstrates the robustness of our method, confirming that these performance trends hold across varied parameter settings. 
    }
    \label{fig:continuous_case_growing_stochasticity_experiments}
\end{figure*}

\subsubsection{Continuous Case Experiments}\label{sec:continuous_case_experiments}
We evaluate \algnamenospace's continuous case method using the same ground robot environment.
Like the discrete case, the robot state is specified with the $x,y$ grid coordinates. 
The robot is restricted to the same discrete action space of up, down, left and right, as in the discrete case. 
For all experiments we set the robot's action step size to $s=0.5$. 
These actions move the robot according to a stochastic transition function, $T$, now characterized by a continuous probability distribution. 
The next state, $T(\state, \action)$, is sampled from a Gaussian centered at the expected transition, $\bar{T}(\state, \action)$, with state and action-dependent covariance, $\Sigma(\state, \action)$. 
The expected transition, $\bar{T}(\state, \action)$, is the deterministic state transition resulting from action $\action$. 
This models the transition as $\bar{T}(\state, \action)$, plus zero-mean Guassian noise. 
To account for the unbounded nature of the Gaussian, in our experiments we truncate noise to within $3$ standard deviations along both $x$ and $y$. 
The \agentspace planner uses the expected transition function, $\bar{T}$, and the \agentspace uses the naive heuristic function based on euclidean distance to the goal center, $x_g$. 

We first validate \algname on an large scale experiment similar to that in the discrete case. 
Here we evaluate both the baseline and our method across the same $8$ ground environments evaluated in the discrete case. 
We run each method across multiple random seeds over a total of $425$ trajectories per method. 
In each experiment the robot is randomly initialized within a start region and must get within radius $r_g$ of the goal region, $x_g$, without exceeding a safety value of $\safetyThreshold=4$. 
The noise variance of each transition in the environment was set to $0.1$ with the algorithm covariance matrix set to $I \times 0.1025$. 

In addition to the above large scale validation, we conduct a second study to evaluate the effect of increasing environmental stochasticity on the performance of our method. 
In these experiments we vary the stochasticity of the environment, by changing the variance of the transitions as well as the variance considered by our algorithm across values of $[0.1, 0.15, 0.175, 0.2, 0.225, 0.25, 0.275, 0.3]$. 
We run each method across the $8$ environments with, $40$ trajectories per environment, per stochastic setting across different random seeds. 
To further test the robustness of our approach we repeat this study across different stochastic configurations across $3$ different GP configurations for each method: config $1$ with lengthscale $1.5$, config $2$ with optimized lengthscales and config $3$ with optimized lengthscales with a slight increase.
The "optimized" GP lengthscales for config $2$ were determined by densely sampling the environment offline and fitting GP kernel hyperparameters, such as the lengthscale, to the data using maximum likelihood. 
The "optimized" GP with increased lengthscales for config $3$ were generated by slightly increasing these "optimized" values. 
The chosen "optimization" procedure, for config $2$, resulted in overly conservative GPs caused by maximizing the likelihood of densely sampled datapoints that shrank the optimized GP lengthscales. 
These parameters lead to the GP sharply increasing its predictive variance with any deviation from the training data. 
The increase in config $3$, relaxes this conservatism. 
Performance across these different configurations is used to highlight effectiveness of our method across suboptimal parameter choices. 

We compare \algname to a baseline method from \cite{safeIML} that does not account for stochasticity. 
Similar to the discrete case we modify the baseline for fair comparison in this stochastic environment, allowing it to check the safety of intended states before transitioning and re-planning if they are unsafe. 
In addition to the standard variants of these methods we also compare to $\beta$ scaling versions of both our method and the baseline, denoted as $\beta_{\text{scaling}} = 0.99$. 
These are variants of our method where the $\beta$ values, which are used by the algorithm to determine pessimistic and optimistic safety, are scaled by $0.99$ every time no target states were found, decreasing them until a minimum permissible $\beta=1.45$. 
This dynamically reduces the conservativeness and safety of the algorithm in favor of performance. 
We compare all methods across the same metrics as the discrete case, success rate, violation rate and number of states explored. 

The results of the first large scale experiment are summarized in Table~\ref{tab:continuous_case_table}. 
These results demonstrate how \algname is able to increase the robot's success rate of reaching the goal while also drastically decreasing the safety violation rate in stochastic environments. 
The lower violation rate is also reflected through the increase in the average number of states explored when using our method. 
The violations and failed attempts of our method primarily arise from the use of sub-optimal GP parameters used to learn the environment's safety costs.  
While higher, static $\beta$ values of the standard algorithm improves safety, this conservativeness comes at the expense of task performance. 
The $\beta_{\text{scaling}}=0.99$ results show how decreasing $\beta$ improves task success by reducing conservativeness at the cost of safety. 
Even with $\beta_{\text{scaling}}=0.99$ our approach is able to achieve a much lower violation rate than the baseline with and without $\beta_{\text{scaling}}=0.99$. 
Additionally, by considering stochasticity and staying safe longer even our standard method is able to outperform the success rate of the baseline with $\beta_{\text{scaling}}=0.99$. 

\begin{figure*}[t]
    \centering
    \begin{subfigure}[t]{0.4\linewidth}
      \centering
      \resizebox{\linewidth}{!}{\includegraphics{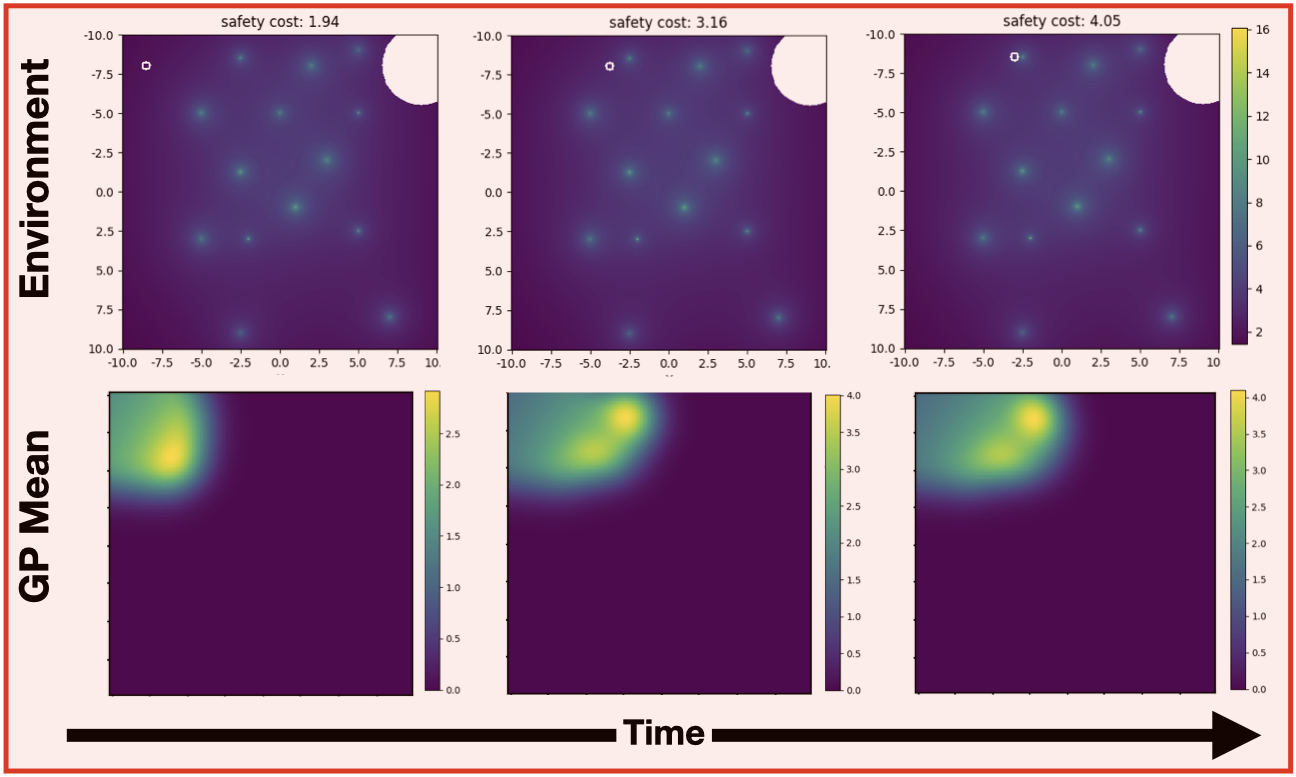}}
      \subcaption{Discrete Case: Baseline}\label{fig:baseline_ground_discrete_traj}
    \end{subfigure}%
    \begin{subfigure}[t]{0.6\linewidth}
      \centering
      \resizebox{\linewidth}{!}{\includegraphics{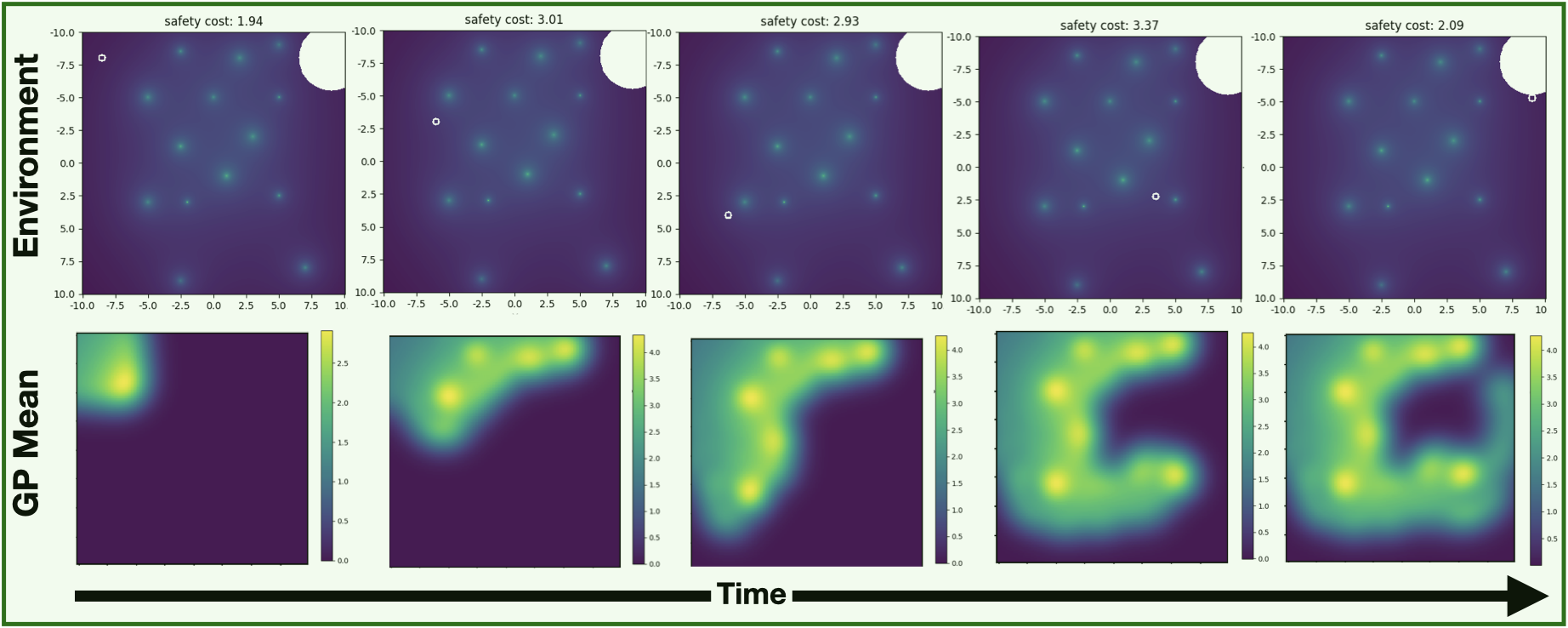}}
      \subcaption{Discrete Case: \algname}\label{fig:ours_ground_discrete_traj}
    \end{subfigure}

    \vspace{0.2em}

    \centering
    \begin{subfigure}[t]{0.385\linewidth}
      \centering
      \resizebox{\linewidth}{!}{\includegraphics{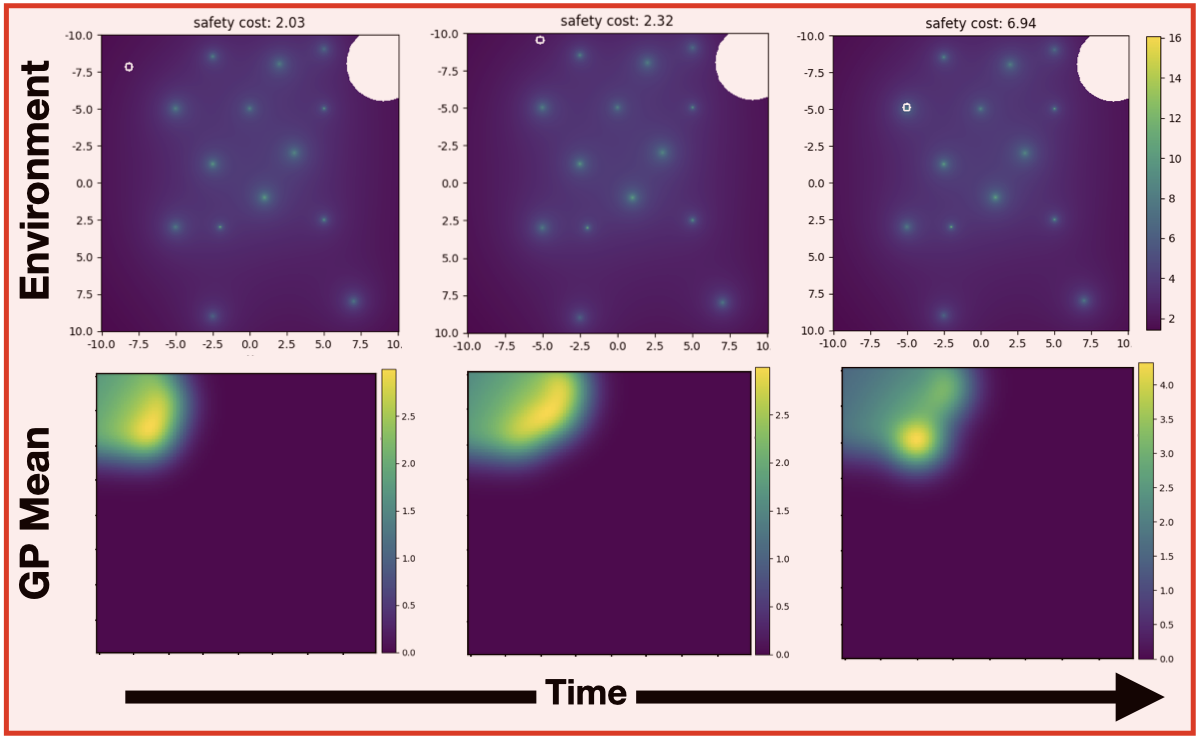}}
      \subcaption{Continuous Case: Baseline}\label{fig:baseline_ground_cont_traj}
    \end{subfigure}%
    \begin{subfigure}[t]{0.6\linewidth}
      \centering
      \resizebox{\linewidth}{!}{\includegraphics{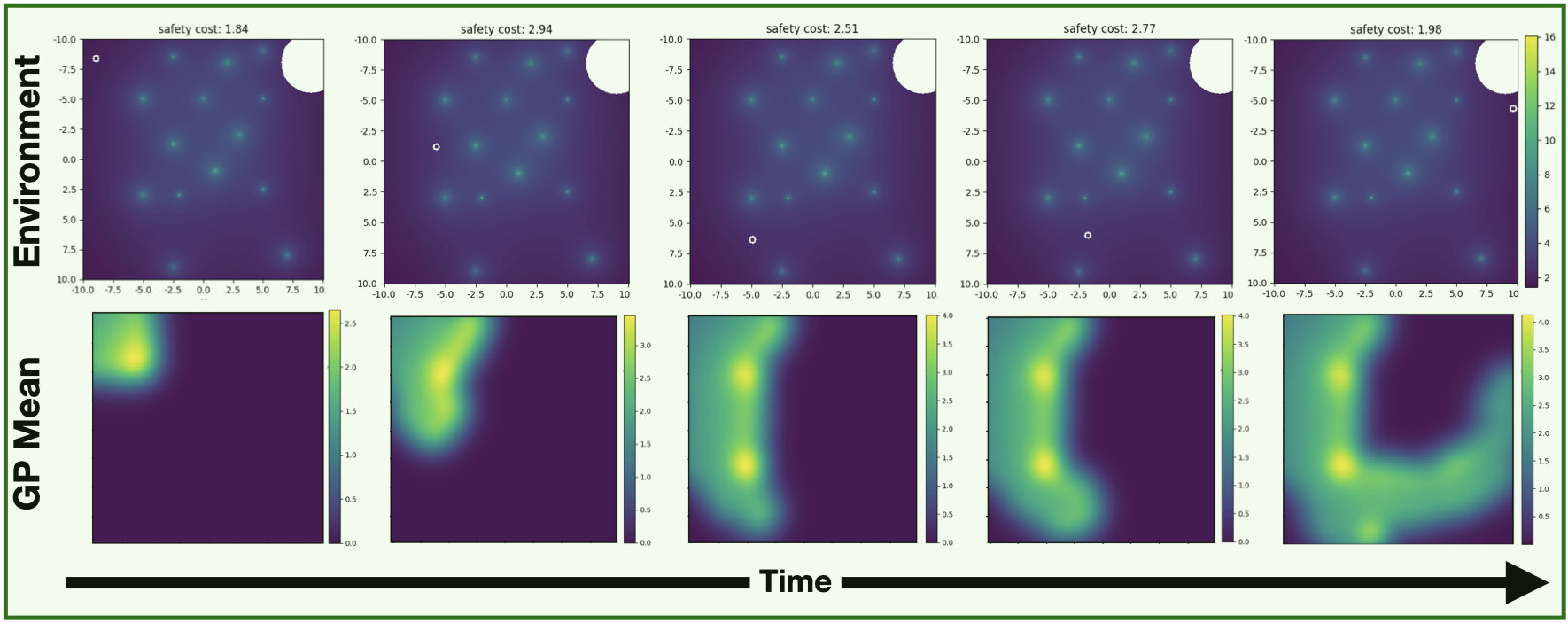}}
      \subcaption{Continuous Case: \algname}\label{fig:ours_ground_cont_traj}
    \end{subfigure}
    
    \caption{
    \textbf{Ground Robot Trajectories: }
    These figures depict snapshots of the ground robot trajectory and corresponding GP predicted mean safety values in top and bottom rows respectively. 
    Sub-figures 
    (a) and (b)
    depict a trajectory for the discrete case experiments (detailed in Section~\ref{sec:discrete_case_experiments}) while 
    (c) and (d)
    depict a trajectory for the continuous case experiments (detailed in Section~\ref{sec:continuous_case_experiments}). 
    Without accounting for stochastic transitions the baseline method quickly violates safety, exceeding the safety threshold of $4$ by entering states with safety values of $4.05$ in the discrete case and $6.94$ in the continuous case respectively. 
    Meanwhile, by accounting for stochasticity our algorithm is able to explore the environment to find a safe path to the goal without violating safety. }
    \label{fig:ground_traj}
\end{figure*}

The results for the second experiment with increasing stochasticity and different GP configurations is shown in Fig.~\ref{fig:continuous_case_growing_stochasticity_experiments}. 
This top plot in this Figure shows the success rate (higher better) of the robot getting to the goal across different stochasticities and GP configurations for each different method. 
The remaining $2$ plots show the safety violation rate (lower better) and the average number of states explored before termination (higher better) due to a success or safety violation for the same set of experiments. 
These experiments demonstrate that as stochasticity increases, \algnamenospace's base method exhibits a slower increase in violation rate compared to the baseline, while maintaining similar success rates. 
The $\beta$ scaling variant of our method performs the best overall. 
By accounting for stochastic transitions, it maintains a lower rate of safety violations than both baseline versions. 
Furthermore, by dynamically balancing safety and performance - reducing conservativeness when progress is no longer possible - our method achieves a higher success rate and also keeps the \agentspace "alive" for longer and enables it to explore significantly ore states than all other approaches. 
Plotting the $3$ GP configurations together showcases the performance ranges of each method even across suboptimal parameters. 
With this, these experiments also demonstrate the robustness of our approach across different GP configurations, where our method is able to maintain a consistently higher success rate and average number of states explored as well as a lower violation rate than both baseline approaches. 
An example trajectory of the $\beta=0.99$ variant of our method in comparison to the baseline is shown in Fig.~\ref{fig:ground_traj}.

\begin{table}[ht!]
\centering
\caption{\textbf{Continuous Case Results: } 
From the large scale Continuous Case Ground Robot experiment in Section~\ref{sec:continuous_case_experiments}. 
}
\label{tab:continuous_case_table}
    \begin{tabular}{lcccc}
    \hline
    \textbf{Method} & \textbf{Successes\%} & \textbf{Violation\%} & \textbf{States Explored} \\
    \hline
    Baseline & 41.8\% & 49.8\% & 171.984 \\
    Baseline $\beta_{\mathrm{scaling}}$ & 48.24\% & 51.29\% & 180.184 \\ 
    \algname & 70.35\% & \textbf{6.82\%} & 266.915 \\ 
    \algname $\beta_{\mathrm{scaling}}$ & \textbf{88.47}\% & 9.88\% & 320.88 \\
    \hline
    \end{tabular}
\label{tab:ground_results}
\end{table}

\begin{figure*}[t]
    \includegraphics[width=\linewidth]{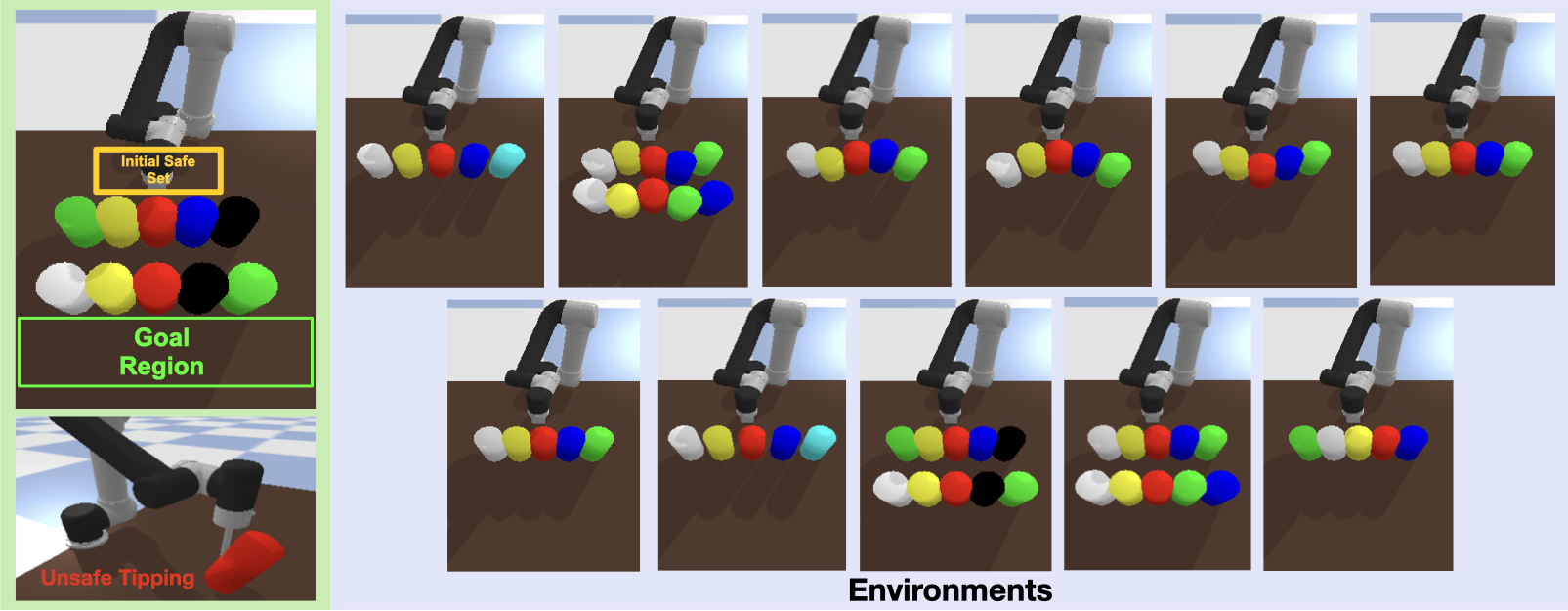}
    \caption{
    \textbf{Safe Object Environments:}
    All $11$ environments used for the Safe Object Experiments. 
    In every environment, the robot is initialized within an initial safe set, having to reach its goal on the other side of the table. 
    The object parameters are varied across each environment and the robot state space is restricted, preventing it from trivially going around the objects. }
    \label{fig:safeobject_envs}
\end{figure*}

\subsection{Safe Object Experiments}\label{sec:safe_object_experiments}
\subsubsection{Safe Object Environment}
As visualized in Fig.~\ref{fig:safeobject_envs}, our second simulated environment is a Safe Object environment. 
Here we simulate a UR5 arm equipped with a flat end-effector attachment to interact with objects arranged in one or two rows, mimicking a typical shelf or pantry configuration. 
The objects are visually identical, but differ in their mass (ranging from $0.25$ to $5$ kg) and center of mass, which alters their stability and interaction dynamics. 
Each object is $20$ cm tall and $5$ cm in radius.

The safety function for object $i$ in the environment is given by: 
\begin{align}
    \label{eq:perobject_safety_function}
    \safetyFunction_{i}(\state_{r}, \state_{o, i}) = \safetyFunction_{\mathrm{tip}, i}(\state_{ro, i}) + \frac{1}{\frac{1}{mc} + d_{i}}
\end{align}
Where $\safetyFunction_{\mathrm{tip}, i}$ is the tipping angle of the object and $\state_{ro, i})$ is the object centric representation of the robot state, $\state_{r}$, with respect to object $i$. 
To ensure a regular change in safety for the GP, the safety cost for each object is determined not only by the object's tipping angle, but also a closeness based cost: $\frac{1}{\frac{1}{mc} + d_{i}}$.
Here $mc=5$ is the maximum closeness cost for the environment and $d_i$ is the robot's distance to to object $i$. 
Safety in the environment is determined by the greatest safety cost across all objects:  
\begin{align}
    \label{eq:objectcase_safety_function}
    \safetyFunction(\state_{r}, [\state_{o, 1}, \dots,\state_{n}]) = \max_{i} \safetyFunction(\state_{r}, \state_{o, i})
\end{align} 
The safety threshold of $23$ is set such that a object tipping beyond $18^\circ$ is considered unsafe. 
The robot's goal is to move its end-effector to the other side, $x \geq 0.75\text{m}$, of the table without exceeding the safety threshold, which could excessively tip or topple over objects, potentially spilling their contents. 
Each environment and the end effector's permissible state space is structured to ensure the that the robot must interact with at least one object to reach the goal. 
The interaction physics of the environment are simulated using PyBullet. 
This environment mimics household, warehouse, or retail scenarios where robots must safely interact with unknown objects.

\begin{table}[ht!]
\centering
\caption{\textbf{Safe Object Results:}
From the Safe Object Experiments in Section~\ref{sec:safe_object_experiments}.
}
\begin{tabular}{lccc}
    \hline 
    \textbf{Method} & \textbf{Success\%} & \textbf{Violation\%} \\ 
    \hline 
    Baseline & 36.36\% & 50.90\% \\ 
    \algname $0.000175$ & \textbf{81.8\%} & 3.64\% \\ 
    \algname $0.0002$ & 76.36\% & \textbf{0\%} \\
    \hline 
\end{tabular}
\label{tab:safe_object_table}
\end{table}

\subsubsection{Object-Centric Case Experiment}
We validate \algnamenospace's object-centric case method
using the Safe Object Environment. 
In our experiments we restrict the motion of the robot end-effector to move in a 2D plane with a specified step size $s=0.01$m using (up $x+=s$, down $x-=s$, left $y-=s$, right $y+=s$) actions  at a fixed height of $z=0.15$m. 
Thus the state space of the robot is entirely defined by the 2D grid coordinates, $x,y$, of the robot's end effector. 
The state space is restricted such that the robot end-effector is unable to trivially go around the objects and such that the end-effector is the only part of the robot allowed to interact with the objects. 
The robot transitions and interactions are simulated using the PyBullet physics engine with transition stochasticity naturally capturing the effects of interacting with unknown objects. 
As in the ground experiments the \agentspace uses a naive expected transition planner that does not account for the object interactions. 
The \agentspace uses a naive heuristic function based on manhattan distance to the goal region of $x\geq 0.75\text{m}$. 

We conducted $55$ random trials across $11$ challenging environments, with distinct configurations of random objects arranged in either $1$ or $2$ rows, as shown in Fig.~\ref{fig:safeobject_envs}. 
All experiments were run across multiple different random seeds, kept identical across methods. 
The random seeds vary the initialization state of the robot end-effector within a specified initial safe set. 
The objective in each run is for the robot end-effector to reach the goal region at the other side of the table without toppling any object beyond the environment's safety threshold value. 

We compare \algname against a deterministic baseline method. 
This baseline is our object-centric case method, but modified to use the baseline method presented in \cite{safeIML} that fails to consider stochasticity. 
This is done for fairness, as without the algorithmic modifications introduced by our object-centric case method, a naive deployment of the baseline method would trivially fail. 
All object-centric methods are run using the continuous case approach to solve for the safe sets. 
We compare our stochastic formulation of the algorithm with transition variances of $0.000175$ and $0.0002$. 
Each method is run with the $\beta$ scaling variant, where the $\beta$ values used by the algorithm to determine immediate safety are autonomously scaled by $0.99$, every time no target states are found, decreasing $\beta$ until a minimum permissible threshold of $\beta=1.45$. 
This slowly relaxes safety and reduces conservativeness in favor of performance. 
We compare all methods across the same metrics as previous experiments, success rate, violation rate and number of states explored. 

\begin{figure*}[t]
    \centering
    \begin{subfigure}[t]{0.49\linewidth}
      \centering
      \resizebox{\linewidth}{!}{\includegraphics{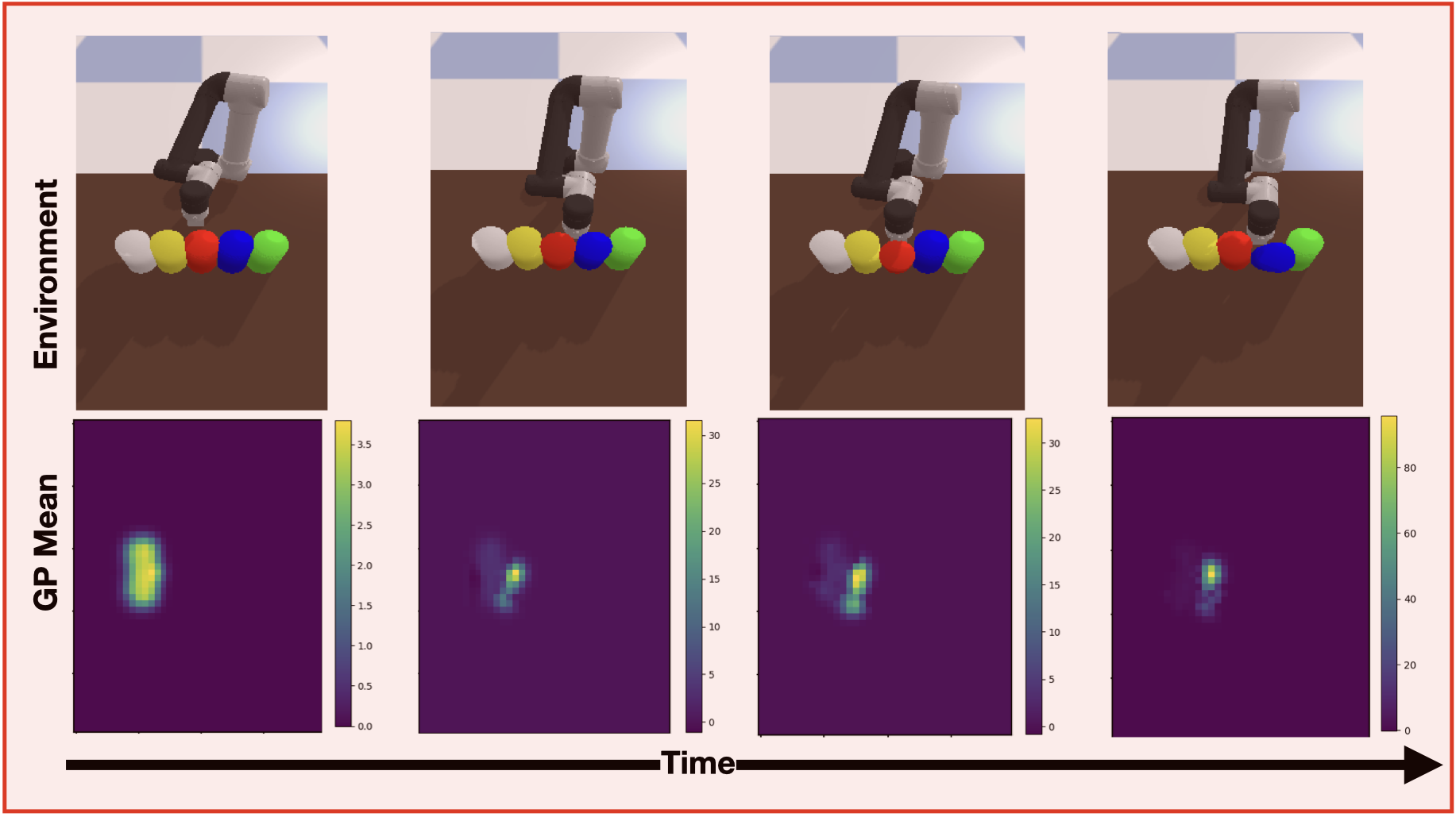}}
      \subcaption{Single Row: Baseline}\label{fig:baseline_safeobject_singlerow}
    \end{subfigure}%
    \begin{subfigure}[t]{0.49\linewidth}
      \centering
      \resizebox{\linewidth}{!}{\includegraphics{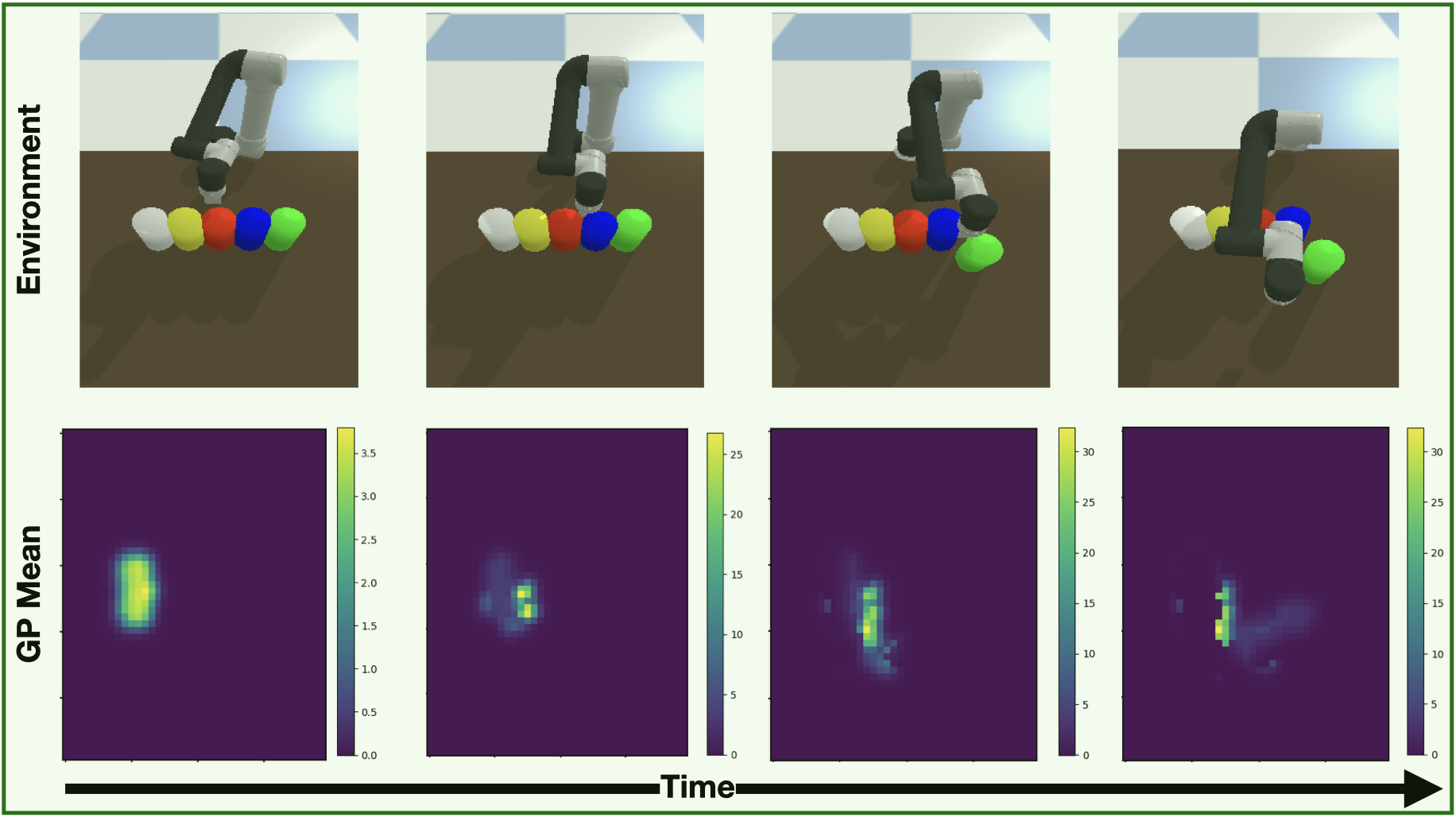}}
      \subcaption{Single Row: \algname}\label{fig:ours_safeobject_singlerow}
    \end{subfigure}

    \vspace{0.2em}

    \centering
    \begin{subfigure}[t]{0.36\linewidth}
      \centering
      \resizebox{\linewidth}{!}{\includegraphics{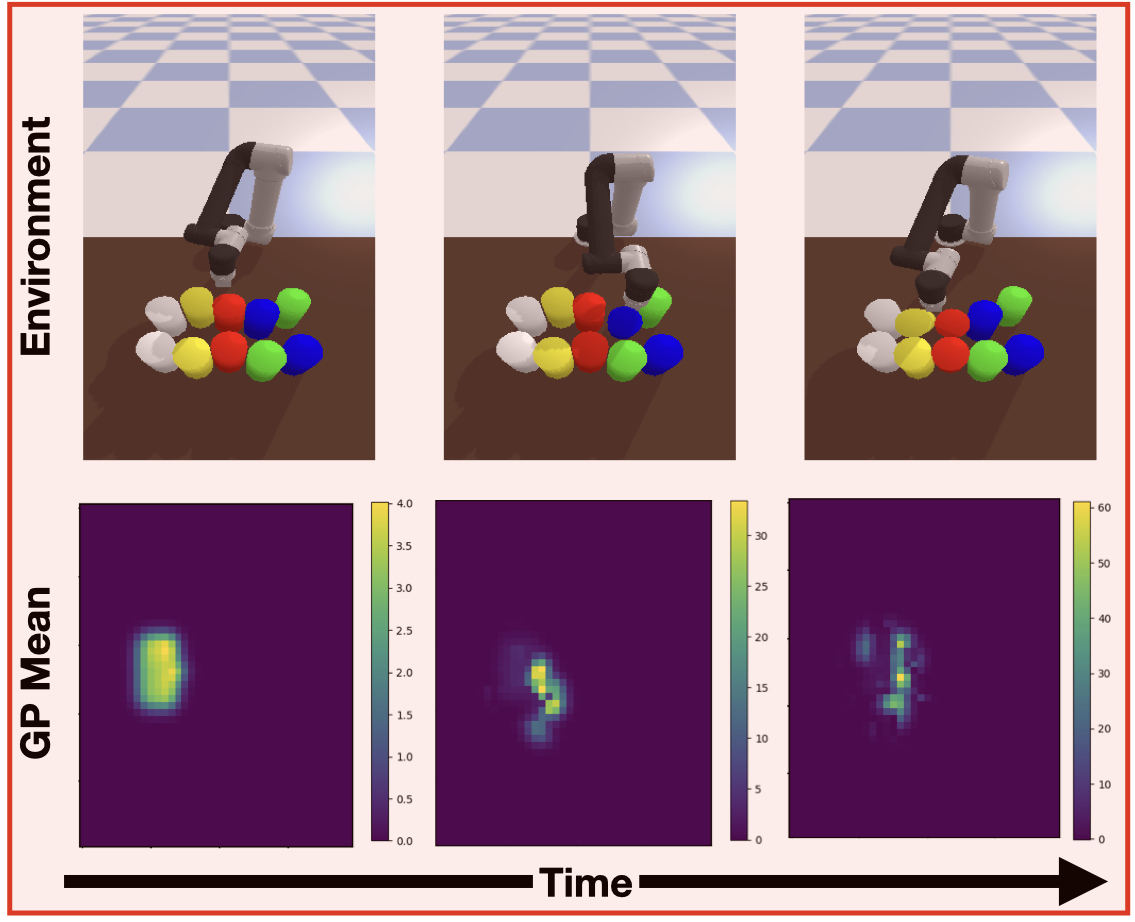}}
      \subcaption{Double Row: Baseline}\label{fig:baseline_safeobject_doublerow}
    \end{subfigure}%
    \begin{subfigure}[t]{0.6\linewidth}
      \centering
      \resizebox{\linewidth}{!}{\includegraphics{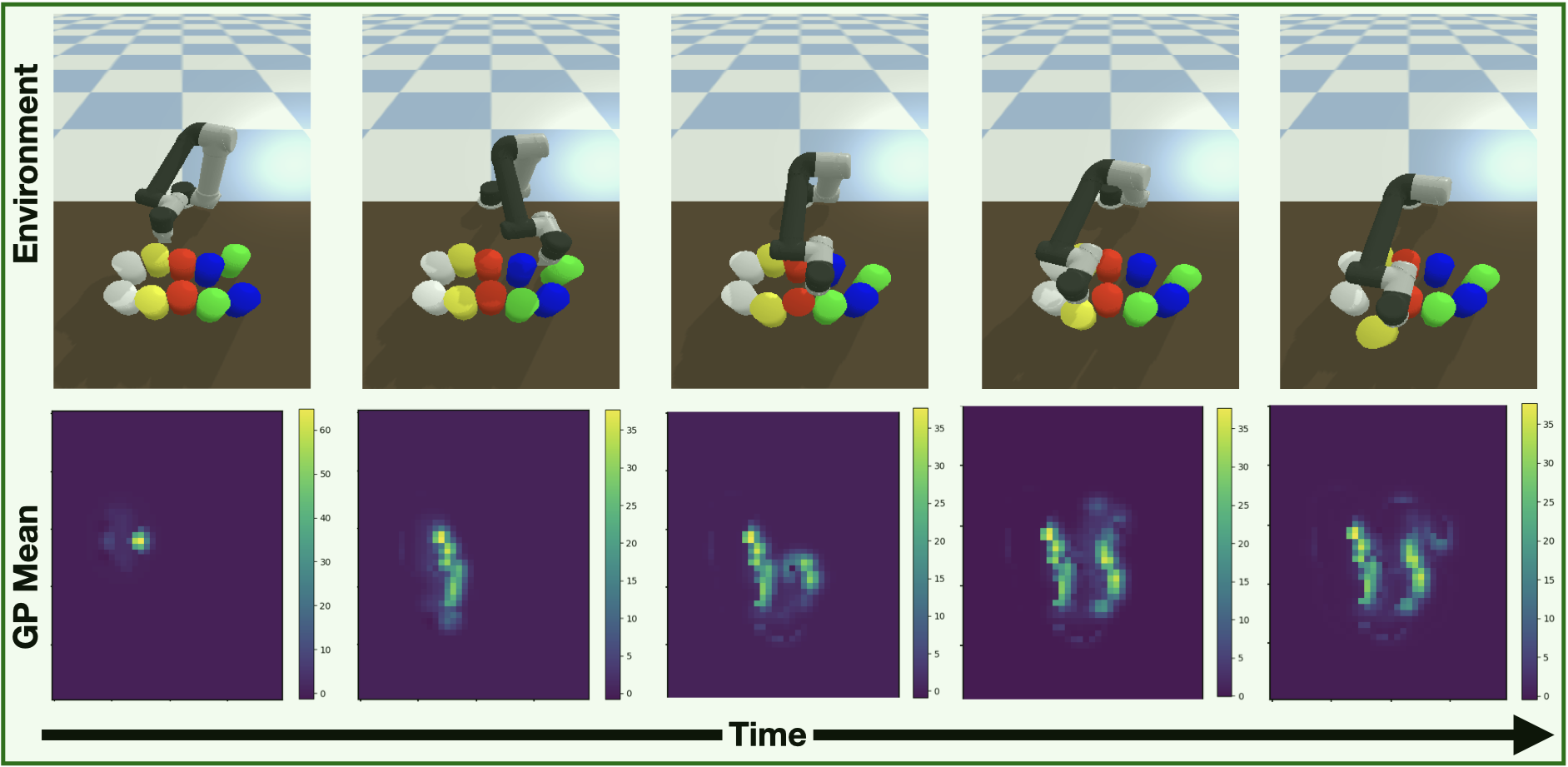}}
      \subcaption{Double Row: \algname}\label{fig:ours_safeobject_doublerow}
    \end{subfigure}
    \caption{
    \textbf{Safe Object Trajectories:} 
    These figures depict snapshots of the safe object robot trajectory and corresponding GP predicted mean safety values from the Safe Object Experiments detailed in Section~\ref{sec:safe_object_experiments}. 
    Sub-figures (a) and (c)
    show trajectories using the baseline method, while (b) and (d)
    show our method on environments with a single and multiple rows of objects respectively. 
    Unable to properly account for the unknown interaction dynamics with different objects, the baseline readily violates safety by excessively tipping over objects.
    Meanwhile, by using stochasticity to account for the unknown interactions our approach is able to safely interact with the objects, finding those that are safe to push out of the way and reach the goal without any safety violations. 
    }
    \label{fig:safe_object_traj}
    \vspace{-0.5cm}
\end{figure*}

The results, summarized in Table~\ref{tab:safe_object_table}, demonstrate that by considering unknown interactions as stochastic transitions \algname reliably reaches the goal region with a higher success rate and decreases unsafe interactions in comparison to the baseline. 
The results from varying stochasticity illustrate that a belief of increased stochasticity improves safety but at the cost of increasing conservativeness. 
Notably, the results also showcase the effectiveness of our object-centric formulation, even enabling the baseline to achieve some safe success. 
Example trajectories of the object case of our method are shown in Fig.~\ref{fig:safe_object_traj}.

\subsection{Hardware Experiments}
\label{sec:hardware_experiments}
Finally we replicate the object case experiments in hardware. 
We use a Franka Emika Panda arm, with a spatula attachment. 
The arm moves in the $x,y$ plane with $1$cm steps. 
The state space of the robot is restricted to prevent trivial trajectories around the objects and ensure interaction. 
Pringles cans, with varied masses and center of mass (achieved by fill them to different degrees with layers of rocks and packing foam) serve as the objects. 
The experimental setup mirrors the simulation, with the arm's end effector starting on one side of a table, trying to reach the other without exceeding the safety threshold, $\safetyThreshold$. 
The robot's state is tracked via inverse kinematics, and the objects' states and tipping angles are measured using using a Kinect camera and ArUco markers. 
We use the same safety function, Eq.~\ref{eq:objectcase_safety_function}, and goal objective as the simulation experiments. 

\begin{figure*}[t]
    \centering
    \begin{subfigure}[t]{0.43\linewidth}
      \centering
      \resizebox{\linewidth}{!}{\includegraphics{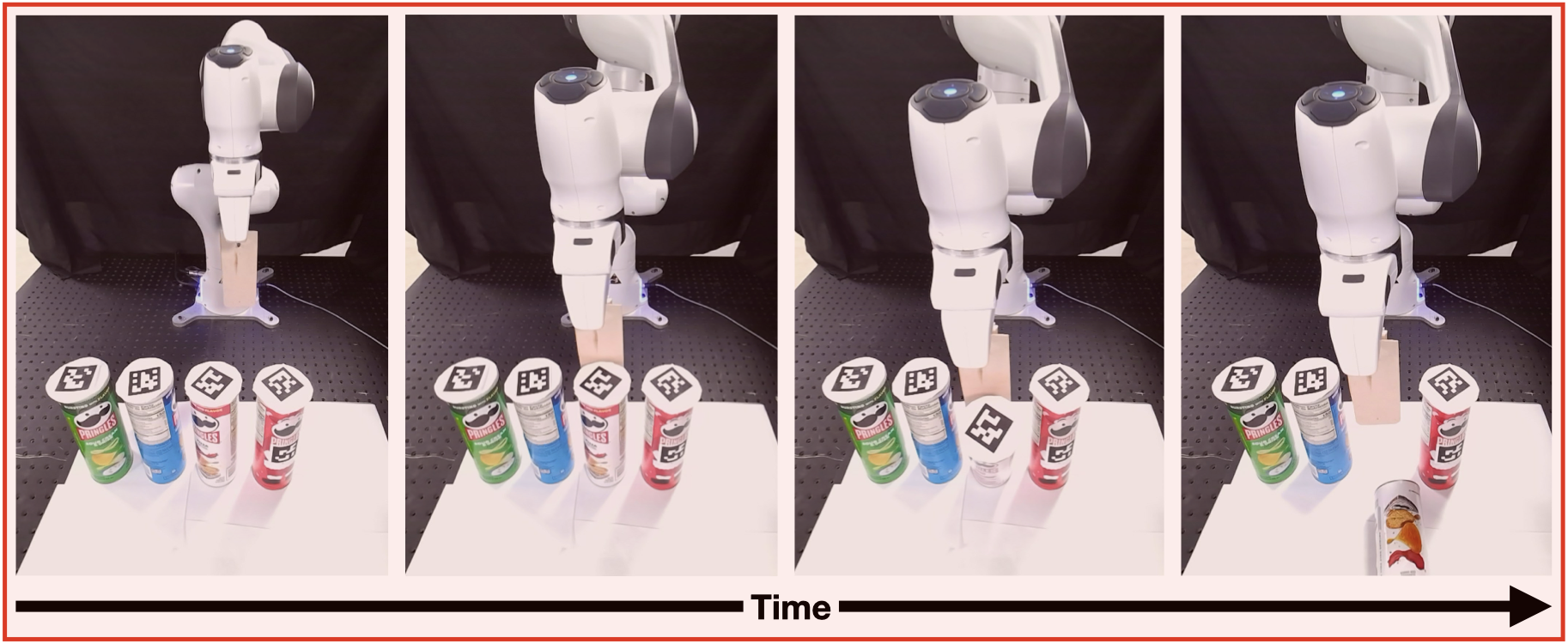}}
      \subcaption{Easy: Baseline}\label{fig:baseline_hardware_easy}
    \end{subfigure}%
    \begin{subfigure}[t]{0.54\linewidth}
      \centering
      \resizebox{\linewidth}{!}{\includegraphics{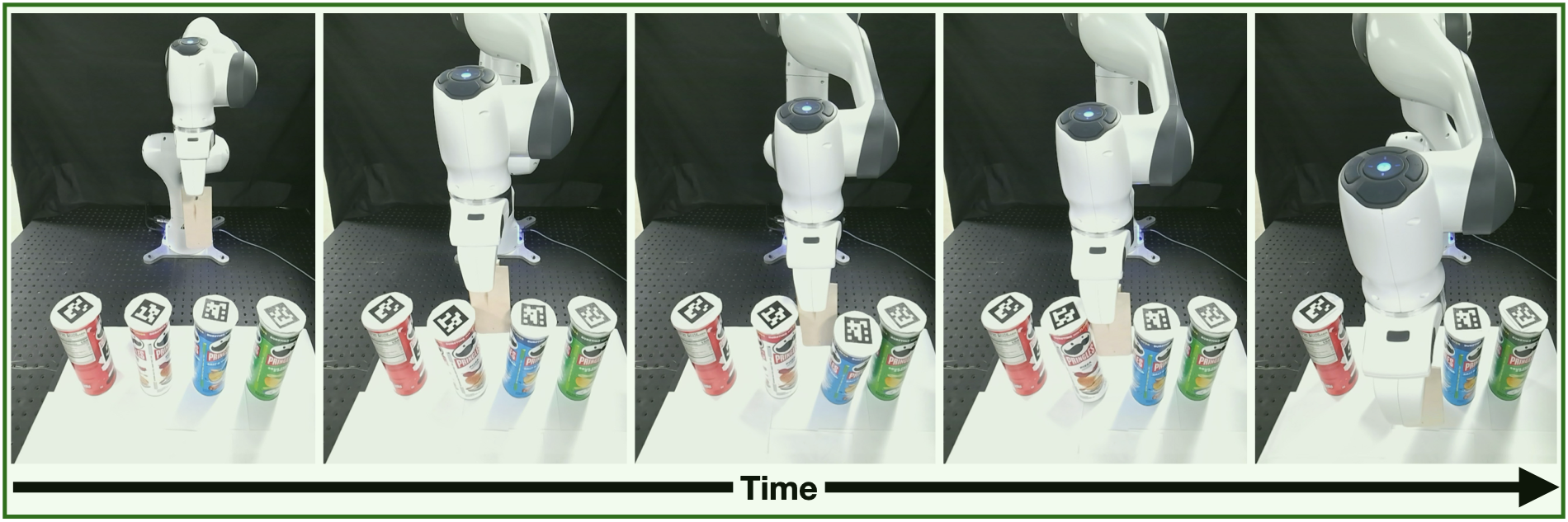}}
      \subcaption{Easy: \algname}\label{fig:ours_hardware_easy}
    \end{subfigure}

    \vspace{0.2em}

    \begin{subfigure}[t]{0.45\linewidth}
      \centering
      \resizebox{\linewidth}{!}{\includegraphics{images/result_trajectory_images/hardware_hard_baseline.png}}
      \subcaption{Hard: Baseline}\label{fig:baseline_hardware_hard}
    \end{subfigure}%
    \begin{subfigure}[t]{0.54\linewidth}
      \centering
      \resizebox{\linewidth}{!}{\includegraphics{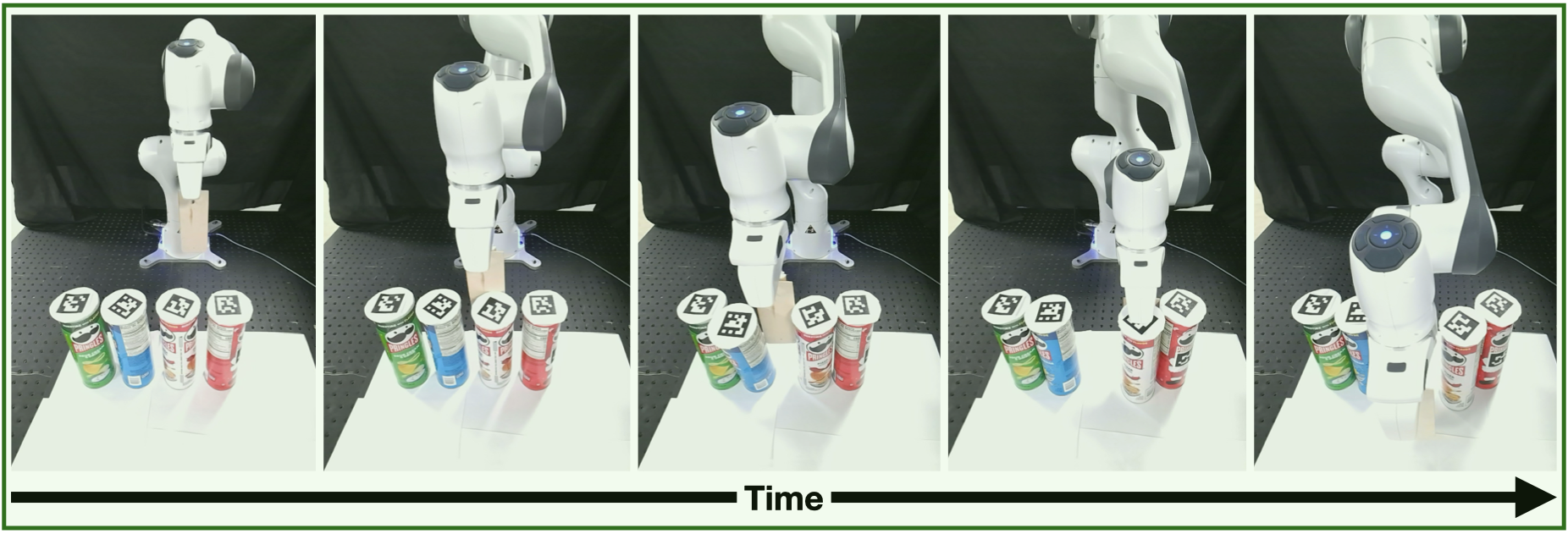}}
      \subcaption{Hard: \algname}\label{fig:ours_hardware_hard}
    \end{subfigure}
    
    \caption{
    \textbf{Hardware Experiment Trajectories}: From experiments in Section~\ref{sec:hardware_experiments}. 
    Sub-figures (a) and (b) show trajectories in an easy environment with more space between objects. 
    Sub-figures (c) and (d) depict trajectories through a hard environment with less space between objects, requiring more robot-object interactions.
    While the baseline quickly violates safety and knocks objects over, our method safely interacts with objects while exploring, allowing the robot to find a safe path to the goal.
    }
    \label{fig:hardware_experiments}
    \vspace{-0.5cm}
\end{figure*}

In hardware we showcase \algname in $2$ different environments, an easy environment shown in Figures~\ref{fig:baseline_hardware_easy} and \ref{fig:ours_hardware_easy} where the objects are placed a little further apart to allow the robot more easily find a path through them, and a hard environment shown in Figures~\ref{fig:baseline_hardware_hard} and \ref{fig:ours_hardware_hard} where the objects are closer together, requiring more interaction. 
In hardware we compare our approach to a naive baseline method that goes directly to the goal without considering the safety of the objects. 
The trajectories of our method and the baseline for the easy and hard environments are shown in Figure~\ref{fig:hardware_experiments}. 
In both cases the robot successfully reached the goal region while keeping all objects within the safety bounds. 
Meanwhile, the naive baseline toppled the objects over, readily violating safety. 
These results demonstrate how \algnamenospace's object-centric case approach readily transfers from simulation to real hardware and enables safe interaction with unknown objects. 

\section{Limitations}
Our work represents significant advances in enabling safe, goal-driven exploration in unknown environments with stochastic transitions. 
While we have introduced novel approaches that extend previous methods to handle stochasticity, as well as taken a first step in enabling safe contact and interaction with unknown objects, our work has limitations that are important to consider and build upon in future work. 

We begin by discussing shared limitations across all variants of our approach. 
One key challenge comes from our reliance of some prior knowledge of the underlying system and its safety function. 
These priors are essential for $3$ aspects of our method: appropriately setting the Lipschitz constant for the discrete case, tuning the GP parameters, and deciding what level of stochasticity to use when modeling the transitions. 
While some knowledge is required, accuracy is not essential. 
Conservative bounds can ensure safety, though at the expense of task performance.
A second shared limitation is the inherent computational scalability of GPs. 
The $O(N^3)$ computational complexity, where $N$ is the number of sampled data points, rapidly increases evaluation costs. 
While GP refinement techniques can mitigate this by culling redundant data points, the computational burden persists in large-scale state spaces.
Finally, all our methods maintain an inherent safety-performance tradeoff. 
Larger $\beta$ values guarantee safety but often induce unnecessary conservativeness.  
While our autonomous and user controlled $\beta$ scaling helps manage this balance, the fundamental tradeoff remains a critical design consideration.

In addition to these shared limitations, each case we consider has its own constraints.

\noindent \textbf{Discrete case: } The discretization scale presents a trade-off. 
Fine discretizations are computationally expensive, whereas excessively coarse grids necessitate a higher Lipschitz constant, which leads to unnecessary conservatism.

\noindent\textbf{Continuous Case: }
As we do not guarantee safe stochastic return, it is possible for the continuous case to get stuck. 
The performance of this method can be heavily impacted by GP kernel tuning: overly large lengthscales risk violating Lipschitz assumptions and inducing unsafe actions, while overly small ones introduce excessive conservatism. 
Furthermore, the core formulation relies on the safety function being adequately represented by a Radial Basis Function (RBF) kernel, although extensions to other analytically tractable kernels are possible.

\noindent\textbf{Object Case: }
This variant also inherits the risk of getting stuck due to the lack of guarantees on safe stochastic return. 
Additionally, the intersection of multiple object-specific safe sets can introduce conservativeness. 
A core limitation is the inability to adequately consider inter-object interactions. 
While extending the GP input to incorporate relative states of nearby objects is theoretically possible, the increased dimensionality would exacerbate the computational constraints introduced by the GP.

\section{Discussion and Conclusion}

In conclusion, we presented \algnamenospace, a method that enables safe, goal-driven exploration in unknown environments with stochastic transitions. 
We extended previous work by proposing and validating novel methods for discrete and continuous state spaces as well as a novel safe object interaction scenario. 
We validated our work extensively in simulations and demonstrated its abilities through hardware experiments. 
Future work will focus on addressing the limitations discussed. 
This includes extending the continuous case to use kernels beyond the RBF kernel, reintroducing the safety condition of safe stochastic return to both the continuous and object cases, and exploring scalable and safe methods to reason about complex inter-object interactions. 

\bibliographystyle{IEEEtran}
\bibliography{bibliography}


\end{document}